\documentclass[wcp]{jmlr}

\jmlrvolume{222}
\editors{Berrin Yan{\i}ko\u{g}lu and Wray Buntine}

\usepackage{longtable}
\usepackage{booktabs}
\usepackage{caption}
\usepackage{lineno}
\usepackage{comment}
\usepackage{float}
\usepackage{graphicx}
\newtheorem{Definition}{Definition}

\newtheorem{innercustomthm}{Lemma}
\newenvironment{mythm}[1]
  {\renewcommand\theinnercustomthm{#1}\innercustomthm}
  {\endinnercustomthm}
\makeatletter
\let\Ginclude@graphics\@org@Ginclude@graphics 
\makeatother

\jmlrvolume{}
\jmlryear{2023}
\jmlrworkshop{ACML 2023}
\jmlrvolume{222}
\title[A Corrected Expected Improvement Acquisition Function Under Noisy Observations]{A Corrected Expected Improvement Acquisition Function Under Noisy Observations}

  \author{\Name{Han Zhou} \Email{han.zhou@esat.kuleuven.be}\\
  \addr Dept. ESAT, Center for Processing Speech and Images,
KU Leuven, Belgium
  \AND
  \Name{Xingchen Ma} \Email{xgchenma@amazon.de\thanks{Work was done at KU Leuven prior to joining Amazon.}} \\
  \addr Amazon Web Services
  \AND
  \Name{Matthew B. Blaschko} \Email{matthew.blaschko@esat.kuleuven.be}\\
  \addr Dept. ESAT, Center for Processing Speech and Images,
KU Leuven, Belgium
 }

\begin{document}

\maketitle

\begin{abstract}
Sequential maximization of expected improvement (EI) is one of the most widely used policies in Bayesian optimization because of its simplicity and ability to handle noisy observations. In particular, the improvement function often uses the best posterior mean as the best incumbent in noisy settings. However, the uncertainty associated with the incumbent solution is often neglected in many analytic EI-type methods: a closed-form acquisition function is derived in the noise-free setting, but then applied to the setting with noisy observations. To address this limitation, we propose a modification of EI that corrects its closed-form expression by incorporating the covariance information provided by the Gaussian Process (GP) model. This acquisition function specializes to the classical noise-free result \citep{jones1998efficient,Mockus1978}, and we argue should replace that formula in Bayesian optimization software packages, tutorials, and textbooks. This enhanced acquisition provides good generality for noisy and noiseless settings. We show that our method achieves a sublinear convergence rate on the cumulative regret bound under heteroscedastic observation noise. Our empirical results demonstrate that our proposed acquisition function can outperform EI in the presence of noisy observations on benchmark functions for black-box optimization, as well as on parameter search for neural network model compression\footnote{The source code is available at \url{https://github.com/han678/correctedNoisyEI}.}.
\end{abstract}


\section{Introduction}
Bayesian optimization (BO) is considered as an effective way to search for a global optimum sequentially, especially when optimizing complex black-box objective function $f(\cdot)$ under limited budgets, that is
\begin{equation}  \small 
       x^{\ast} = \arg \max_{x \in\mathcal{X}} f(x) 
\end{equation}
where $\mathcal{X} \subseteq \mathbb{R}^d$ represents the bounded input space. Its optimization procedure relies on a Gaussian Process (GP) model that allows us to relax the assumption of the objective functions, leading to its popularity in many important applications including experimental particle physics, material design, and hyper-parameter tuning for machine learning algorithms. Once we have the GP surrogate model, the sequential selection of BO can be made through a decision function, known as the acquisition function that relies on this model. Among existing acquisition functions, Expected Improvement (EI) is one of the most widely used as it preserves a good balance between exploration and exploitation. It is generally defined as the expectation of the improvement function at iteration $t$: $\mathbb{E}[I_t(x)]$ where the improvement function $I_t(x)=\max\{0,f(x)-\xi\}$ relies on the incumbent solution $\xi$. Let us denote the set of sampled points up to iteration $t-1$ as $\mathcal{D}_{t-1}$, then in the noiseless setting where $f(x)$ can be easily observed, the incumbent best at iteration $t$ is thus given by $\xi = f(x_t^+)$ with the best point $x_t^+ = \arg\max_{i\leq t-1}f(x_i)$. However, in the framework of noisy observations i.e. $y_t=f(x_t)+\varepsilon_t$, the true value of the objective function is not exactly known due to the noise term $\varepsilon_t$ on the observations. Several existing works consider a plug-in estimate as the incumbent best to tailor the improvement function for the noisy environment. Popular substitutes include the best \emph{noisy} observation $\max_{i\le t-1}y_i$ \citep{nguyen2017regret} and the best value of the GP predictive mean $\arg\max\mu_{t-1}(x_i)$ over the input space $\mathcal{X}$~\citep{wang2014theoretical} or the observation set $\mathcal{D}_{t-1}$~\citep{vazquez2008global, scott2011correlated}. Then the expectation of those improvement functions can be calculated as that in the noiseless case. 

\begin{figure}[H]
\centering
\includegraphics[width=5in]{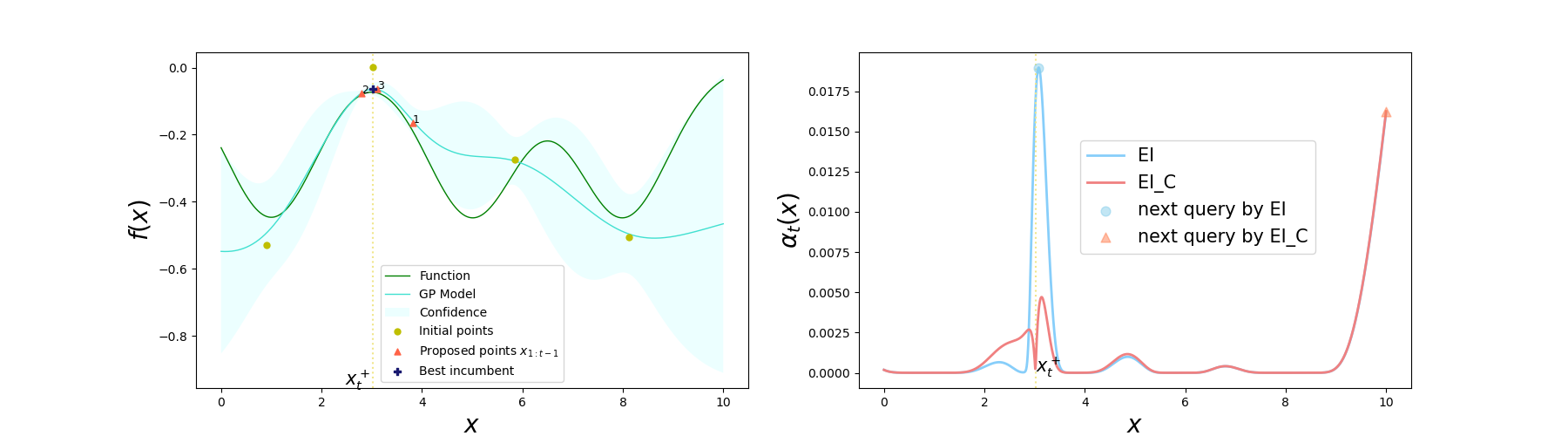}
\caption{A scenario example illustrates that the utilization of covariance information has the potential to circumvent local search in a noisy environment. The goal is to locate the maximum of this objective function. Notably, the corrected EI suggests a point that is closer to the global optimizer $x^{\ast}=10$.}
\label{fig:uncertainty}
\end{figure}

 However, existing analytic EI-type acquisitions~\citep {gupta2022regret} often treat that plug-in estimate as deterministic throughout each iteration and do not consider its uncertainty when formulating their closed-form expressions, which potentially leads to local search behavior in some circumstances (see Figure~\ref{fig:uncertainty}). Apart from that, noticing the incumbent solution comes with uncertainty, then depending on the type of selected kernel, the covariance information between $x_t^+$ and other points may need to be considered as well when specifying its analytic expression. For example, if we specify our GP model with a white noise kernel,\footnote{In this case, BO degenerates to random search.} then this issue can be ignored since this kernel simply assumes all covariances between samples to be zero. However, this assumption does not hold for widely used kernels such as the Matérn 
 and Squared Exponential kernels, especially when the length scale parameter is large. To address these challenges, we introduce a novel acquisition function that effectively incorporates the uncertainty of the incumbent solution. We consider an improvement function with an unknown objective value over the best point that maximizes the GP predictive mean as the best incumbent, akin to the noiseless scenario. Although lacking a deterministic incumbent in our improvement function, we derive an analytical expression for this acquisition function in the presence of noisy observations, which also generalizes that of the noiseless case. Our acquisition is constructed directly from the correct variance, allowing us to take full advantage of the covariance information from the GP model and generate its closed-form representation under noisy observations. Furthermore, we provide a regret bound for this acquisition function under heteroscedastic observation noise. The effectiveness of our method is further demonstrated through our empirical experiments.
\section{Related work}
For a recent review of Bayesian optimization, we refer interested readers to \citet{garnettbayesoptbook2022}. The expected improvement under noisy data is more complicated than in the noiseless case~\citep{jones1998efficient}, due to the uncertainty in the noisy observations. \citet{santner2003design} stressed the importance of measuring the noise variance values and taking them into account when fitting the GP model. In the noisy setting, the best value of the objective function is unknown, thus substituting a plug-in estimate is popular in practice, such as the maximum noisy observation \citep{nguyen2017regret}. However, this estimate lacks robustness as it may have high variance \citep{picheny2013benchmark}. An alternative choice is the best GP predictive mean $\max_{x\in\mathcal{S}} \mu_{t-1}(x)$ where $\mathcal{S}$ can be the search space $\mathcal{X}$ \citep{wang2014theoretical} or the observation set $D_{t-1}$ \citep{vazquez2008global, gupta2022regret}. The additional computation for the predictive mean adds more cost to the optimization, especially for a large search space $\mathcal{X}$, but may be acceptable for optimizing functions that are very expensive to evaluate. With these plug-in estimators, the noiseless EI acquisition function can be applied also when there are noisy observations. There have been several works that investigate the convergence properties of the regret bound for EI and improved EI-type algorithms. \citet{bull2011convergence} established an upper bound of the simple regret for EI in the noiseless scenario. \cite{wang2014theoretical} derived a regret bound for expected improvement with the best predictive mean of the GP model as the incumbent best under the noisy setting. \cite{nguyen2017regret} proved a regret bound for EI with the best-observed value as the incumbent best. Their regret bound relies on a pre-defined termination threshold $\kappa$, which is also assumed in this paper. 

Apart from plug-in estimates, \cite{forrester2006design} provided an approximation for EI through re-interpolation, which relies on a noise-free GP using the predicted data made by the noisy GP model. More rigorous approaches estimate EI through Monte Carlo (MC) integration \citep{williams2000sequential, letham2019constrained, balandat2020botorch}, which results in more expensive computation. In particular, \cite{letham2019constrained} proposed an MC-based EI-type acquisition that handles the uncertainty by averaging over the EI values of a number of noisy-free GP models. On the other hand, \cite{balandat2020botorch} introduced another MC-based acquisition that considers the uncertainty stemming from the unknown incumbent best. They address this uncertainty by averaging the improvement functions on a set of $q$ test points and previously evaluated points, which allows for parallel (batch-sequential) optimization to accelerate the search process. These methods nevertheless remain significantly more computationally expensive than closed-form acquisition functions.

There are several other popular acquisition functions including probability of improvement (PI) \citep{kushner1964new}, upper confidence bound (UCB), knowledge gradient (KG), entropy search (ES), and predictive entropy search (PES) \citep{hernandez2014predictive}. The last three methods are most useful in exotic problems where the assumptions made by EI does hold anymore. Similar to EI, the PI acquisition prefers selecting points near the incumbent best which can potentially result in over-exploitation~\citep{brochu2010tutorial}. Recent studies by \citet{ma2019bayesian} combine PI with covariance information from the GP model under a noisy environment. On the other hand, the UCB acquisition function emphasizes exploring areas with higher uncertainty, promoting more exploration. \cite{srinivas2010gaussian} established a regret bound for UCB in the noisy setting using the information capacity $\gamma_i$ and a parameter $\beta_i$ that depends on the reproducing kernel Hilbert space (RKHS).

The main contribution of this paper is a modified expected improvement that leverages the covariance information from the heteroscedastic GP model \citep{le2005heteroscedastic}, which is more applicable to noisy environments. In this model, the input measurements are treated as deterministic, leading to the noise variance varying across the input space and allowing us to utilize more precise prior covariance information for noisy observations. To our knowledge, we provide the first corrected expected improvement that directly incorporates the uncertainty of the incumbent solution under noisy observations, along with an upper bound on the regret associated with this acquisition.

\section{Bayesian optimization}
As mentioned before, Bayesian optimization aims to find the global optimum of a black-box function $f(\cdot)$ on a bounded input space $\mathcal{X}$. Drawing noisy samples from the objective function is typically expensive, making it essential to enhance sampling efficiency. BO tackles this challenge by utilizing a GP surrogate model that could also help balance the exploration-exploitation trade-off during the search process. 

\subsection{Modelling with Gaussian processes}
A Gaussian process inherits the elegant mathematical properties of the multivariate normal distribution and provides a flexible framework for modeling the objective function $f(\cdot)$. Typically, the model specifies $f(x)$ as a Gaussian process $\mathcal{GP}\left(m(x), k(x_i,x_j)\right)$ with a mean function $m(x)$ and a positive semi-definite covariance matrix (or kernel) $k(x_i,x_j)$. In the presence of observation noise $\varepsilon_i$, we observe $y_i = f (x_i) + \varepsilon_i$ instead of the objective value. The noise term is assumed to be $\varepsilon_i \sim \mathcal{N}(0,\upsilon_i^2 )$ for the purposes of GP regression. Let $\mathcal{D}_{t-1}=\{(x_i,y_i, \upsilon_i)\}_{i=1}^{t-1}$ be the set of noisy observations with uncertainty estimates up to iteration $(t-1)$ of Bayesian optimization. Assuming a prior distribution $\mathcal{GP}\left(0, k(x_i,x_j)\right)$ over $f$, similar to the noiseless case, the posterior distribution also follows a Gaussian distribution $P(f|\mathcal{D}_{t-1})=\mathcal{GP}(\mu_{t}(x),\sigma_{t}^2(x))$ where 
\begin{equation} \label{eq:approximationGp}  \small
\begin{aligned}
\mu_{t}(x) &= k_{t-1}(x)^T\left(K_{t-1}+ \Sigma_{t-1} \right)^{-1} \boldsymbol{y}_{t-1}\\
 \sigma_{t}^2(x) &= k_{t-1}(x,x)- k_{t-1}(x) \left(K_{t-1}+ \Sigma_{t-1}\right)^{-1} k_{t-1}(x)^T  
\end{aligned}
\end{equation}
with $\boldsymbol{y}_{t-1}= [y_1, \cdots, y_{t-1}] $, covariance matrix $K_{t-1}=[k_{t-1}(x_i,x_j)]_{1\le i,j\le t-1}$ and $k_{t-1}(x) = [k_{t-1}(x_1,x),\cdots,k_{t-1}(x_{t-1},x)]$. Here $\Sigma_{t-1} = diag(\upsilon_1^2, \cdots,\upsilon_{t-1}^2)$ is a diagonal matrix formed by the variances of the noise terms. In addition, the covariance between $x_i$ and $x_j$ is 
\begin{equation}  \small
\begin{aligned}
    \sigma_{t-1}(x_ix_j) = k_{t-1}(x_i, x_j)-k_{t-1}(x_i) \left(K_{t-1}+ \Sigma_{t-1}\right)^{-1} k_{t-1}(x_j)^{T} 
\end{aligned}
\end{equation} 
which indicates the relationship between these two points. As such, the shape (smoothness, amplitude of the predictive variance) of the GP model is significantly influenced by the choice of the kernel as well as the kernel parameters. A large number of covariance kernels are available in the literature \citep{santner2003design}. The Matérn kernel, as one of the most widely used kernels, is defined as 
\begin{equation}  \small
k_{\text {Matérn}}\left(x_i, x_j\right)=\frac{2^{1-\nu}}{\Gamma(\nu)}\left(\frac{\left\|x_i-x_j\right\|_2}{\ell}\right)^\nu \mathcal{B}_\nu\left(\frac{\left\|x_i-x_j\right\|_2}{\ell}\right)
\end{equation}
where $\mathcal{B}_\nu$ is the modified Bessel function, $\Gamma(\cdot)$ is the Gamma function, $\nu$ is the smoothness parameter, and $\ell$ is the scale parameter. A popular Matérn kernel is Matérn-$\frac{5}{2}$ with  $\nu=\frac{5}{2}$, which satisfies the twice differentiable property. Another popular kernel is the square exponential kernel, which is given by $k_{se}(x_i, x_j)=\exp (-\frac{\| x_i-x_j\|^{2}}{2l^{2}})$ with length scale $\ell$. 

\subsection{Expected Improvement}
Once we have the GP model built on the observation set, we can employ BO alongside an appropriate acquisition function $\mathcal{\alpha}_t(x)$ to identify the next point $x_t$ for evaluation. This candidate is obtained by maximizing the acquisition function, that is $x_t=\arg\max_{x\in \mathcal{X}} \mathcal{\alpha}_t(x)$. 
The expected improvement balances exploration and exploitation by maximizing the expectation over the improvement function $I_t(x)$. Some authors \citep{movckus1975bayesian,lizotte2008practical} have introduced an additional parameter $\zeta$ to augment this criterion as $I_t(x)=\max\{0, f(x)-\xi-\zeta\}$, but this specific augmentation is out of the scope of this paper and will not be further discussed. In this paper, we consider the improvement function $I_t(x)=\max\{ 0,f(x)-\mu_{t-1}(x_t^{+})\}$ with  $x_t^{+}=\arg\max_{i\le t-1}\mu_{t-1}(x_i)$. The incumbent best is defined as the best GP predictive mean over the observation set. This acquisition function can be evaluated in closed form as~\citep{wang2014theoretical}:
\begin{equation}  \small
\begin{aligned}
\mathcal{\alpha}_t(x) = \mathbb{E}\left[ \max\{ 0,f(x)-\mu_{t-1}(x_t^{+})\} \right]=\sigma_{t-1}(x) \phi(z) + \left(\mu_{t-1}(x)- \mu_{t-1}(x_t^{+})\right)\Phi(z)
\end{aligned}
\end{equation}
where $z = \left( \mu_{t-1}(x)- \mu_{t-1}(x_t^{+})\right)/ \sigma_{t-1}(x)$, $\phi$ is the standard normal PDF, and $\Phi$ is the standard normal CDF. The computation cost of this acquisition function is far cheaper than the black box function. 
\section{Corrected Expected improvement}
 Our modified expectation improvement aims to maximize the expectation of the improvement function, denoted as $I_t^{C}(x)=\max\{0,f(x)-f(x_t^+)\}$, s.t.
\begin{equation}  \small
\mathcal{\alpha}^{C}_t(x)= \mathbb{E}(\max\{0, f(x)-f(x_t^+)\})  \label{eq:MEI}
\end{equation}
where $x_t^+ = \arg\max_{i\le t-1}\mu_{t-1}(x_i)$. We note that the incumbent best $f(x_t^+)$ is defined as the unknown true objective value of the best point among the observation set that maximizes the predictive mean of the GP model. Let us define $\Tilde{\sigma}_{t-1}^2(x)= \sigma_{t-1}^2(x)+ \sigma_{t-1}^2(x_t^+)-2\sigma_{t-1}(xx_t^+)$ where $\sigma_{t-1}^2(\cdot)$ is its corresponding variance, and $\sigma_{t-1}(xx_t^+)$ is the covariance between points $x$ and $x_t^+$. Despite not knowing the incumbent best in our improvement function, we can still derive a closed-form expression for our acquisition in noisy settings (Section \ref{Appendix:derivativeMEI}):
\begin{equation}  \small
  \mathcal{\alpha}^{C}_t(x) =  \Tilde{\sigma}_{t-1}(x) \phi\left(\frac{\mu_{t-1}(x)-\mu_{t-1}(x_t^+)}{\Tilde{\sigma}_{t-1}(x)}\right) + \left(\mu_{t-1}(x)-\mu_{t-1}(x_t^+)\right)\Phi\left(\frac{\mu_{t-1}(x)-\mu_{t-1}(x_t^+)}{\Tilde{\sigma}_{t-1}(x)}\right)
\end{equation}
where $\phi$ and $\Phi$ are the density and cumulative distribution functions of the standard normal distribution, respectively. When $\Tilde{\sigma}_{t-1}(x)=0$, we set $\mathcal{\alpha}^{C}_t(x) =0$. In the noiseless case where $\forall x\neq x_t^+\text{, } \sigma_{t-1}(x_t^+) = \sigma_{t-1}(xx_t^+) = 0 $, we recover the expression for EI. In addition, let us define the function $z_{t-1}(x)=\frac{\mu_{t-1}(x)-\mu_{t-1}(x_t^+)}{\Tilde{\sigma}_{t-1}(x)}$ and the function $\tau(z)=z\Phi(z)+\phi(z)$, then we derive another expression of this acquisition function
\begin{equation}  \small
  \mathcal{\alpha}^{C}_t(x) = \Tilde{\sigma}_{t-1}(x)\tau(z_{t-1}(x))\label{def:MEI2}
\end{equation}
where $\Tilde{\sigma}_{t-1}(x)$ is non-negative and it reaches zero at $x_t^+$ even in the noisy setting. From this expression, the corrected variance term $\Tilde{\sigma}_{t-1}(x)$ tends to zero when $x\rightarrow x_t^+$. As a result, the corrected expected improvement of points next to the current best point would be relatively small, avoiding over-exploration around this region. It makes intuitive sense that this acquisition function would search more globally than EI. An example in Figure~\ref{fig:uncertainty} illustrates this fact. 

\subsection{Derivation of the Modified Expected Improvement} \label{Appendix:derivativeMEI}
When $f(x)-f(x_t^+)$ is non-negative, the variable $I_t^{C}(x)$ is Gaussian distributed with mean $u_{t-1}(x) = \mu_{t-1}(x)-\mu_{t-1}(x_t^+)$ and variance $\Tilde{\sigma}_{t-1}^2(x)= \sigma_{t-1}^2(x)+ \sigma_{t-1}^2(x_t^+)-2\sigma_{t-1}(xx_t^+)$ where $\mu_{t-1}(x)$ is the mean of the GP evaluated at $x$, $\sigma_{t-1}^2(x)$ is its corresponding variance at $x$, and $\sigma_{t-1}(xx_t^+)$ is the covariance between points $x$ and $x_t^+$. Thus using the likelihood of $I_t^{C}(x)$ (for simplicity, we write $I$), we obtain the expectation of our improvement function:
\begin{equation}  \small
\begin{aligned}
\mathcal{\alpha}_t^{C}(x)&= \int_{0}^{\infty}  \frac{I}{\sqrt{2\pi}\Tilde{\sigma}_t(x)}\exp\left(-\frac{1}{2}\left(\frac{I-u_{t-1}(x)}{\Tilde{\sigma}_{t-1}(x)}\right)^2\right)dI \label{eq:likelihood}
\end{aligned}
\end{equation}
Let $s=\frac{I-u_{t-1}(x)}{\Tilde{\sigma}_{t-1}(x)}$, then $I=s\Tilde{\sigma}_{t-1}(x)+u_{t-1}(x)$ and $ds = \frac{1}{\Tilde{\sigma}_{t-1}(x)}dI$. Using the above likelihood \eqref{eq:likelihood}, we obtain our modified expected improvement 
\begin{equation}  \small
\begin{aligned}
\mathcal{\alpha}_t^{C}(x) & = \int_{-\frac{u_{t-1}(x)}{\Tilde{\sigma}_{t-1}(x)}}^{\infty} \frac{s\Tilde{\sigma}_{t-1}(x)+u_{t-1}(x)}{\sqrt{2\pi}\Tilde{\sigma}_{t-1}(x)}\exp\left(-\frac{s^2}{2}\right)\Tilde{\sigma}_{t-1}(x) ds\\
 &= \frac{\Tilde{\sigma}_{t-1}(x)}{\sqrt{2\pi}}\int_{-\frac{u_{t-1}(x)}{\Tilde{\sigma}_{t-1}(x)}}^{\infty}s e^{-\frac{s^2}{2}}ds +u_{t-1}(x) \int_{-\frac{u_{t-1}(x)}{\Tilde{\sigma}_{t-1}(x)}}^{\infty}\frac{e^{-\frac{s^2}{2}}}{\sqrt{2\pi}}ds \\
&= \frac{\Tilde{\sigma}_{t-1}(x)}{\sqrt{2\pi}}(-e^{-\frac{s^2}{2}})|_{-\frac{u_{t-1}(x)}{\Tilde{\sigma}_{t-1}(x)}}^{\infty} +u_{t-1}(x) \Phi\left(\frac{u_{t-1}(x)}{\Tilde{\sigma}_{t-1}(x)}\right)\\
&=\Tilde{\sigma}_{t-1}(x) \phi\left(\frac{u_{t-1}(x)}{\Tilde{\sigma}_{t-1}(x)}\right) + u_{t-1}(x) \Phi\left(\frac{u_{t-1}(x)}{\Tilde{\sigma}_{t-1}(x)}\right)  . \label{eq:modified_ei}
\end{aligned}
\end{equation}

\subsection{Termination Criterion}
Similar to \citet[Lemma~2]{nguyen2017regret}, the value of our acquisition function is set to be greater than a positive value $\kappa$ that is $ \max_{t} \alpha^{C}(x_t) \geq \kappa$. This termination criterion guarantees the convergence of the regret bound for our acquisition function. We will utilize this property to show our regret bound in Section~\ref{sec:theoreticalpart}.  We additionally show here a connection to classical results in economics, thereby giving an interpretation of the value of $\kappa$ and its scale relative to $f(\cdot)$.

\subsubsection{Profit-Cost curve} 
\begin{figure}[h]\footnotesize
\centering
\includegraphics[width=1.6in]{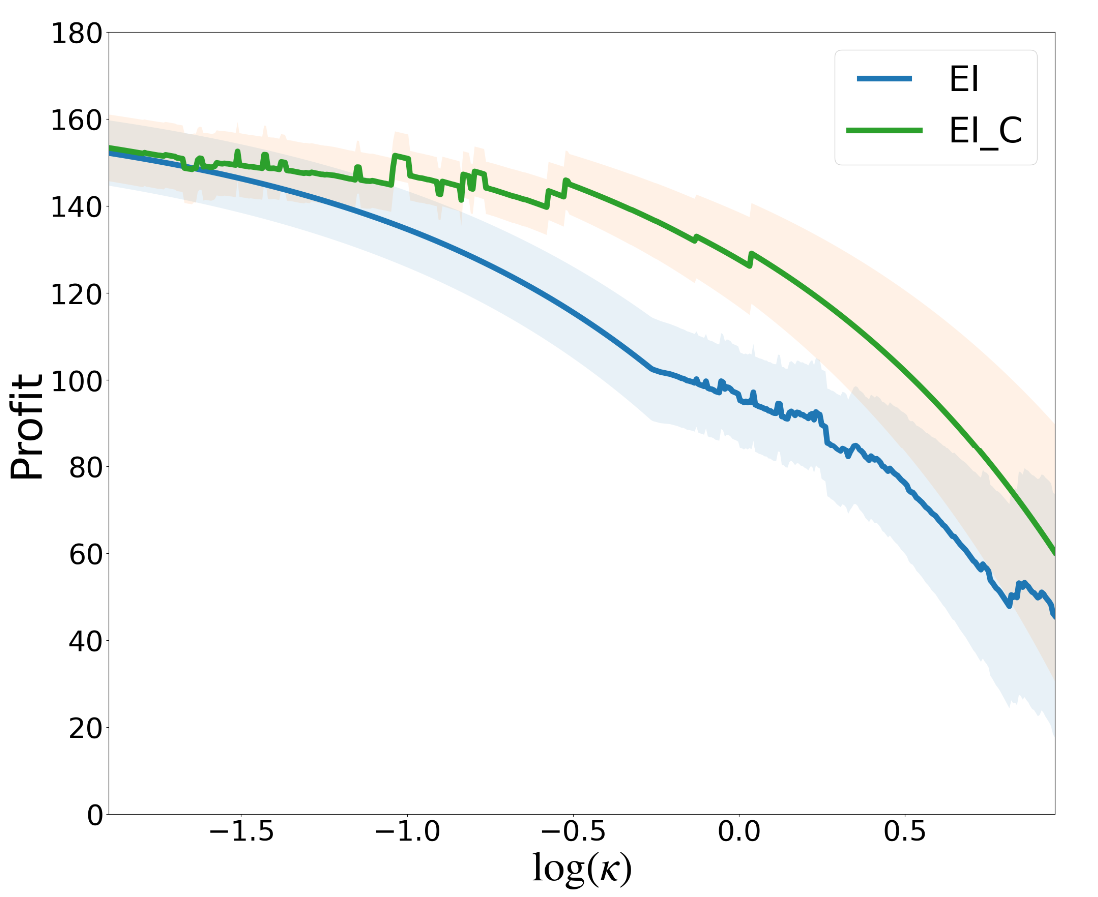}
\caption{Comparison of EI (blue line) and corrected EI (green line) over the profit for same termination criterion $\kappa$ on the \textit{Sphere 3d} function. In our simulations, the standard deviation of the Gaussian noise term is set to be 20. The average profit is measured over 15 experiments for each $\kappa$.}
\label{fig:profitcurve}
\end{figure}

Let us denote $t_{\kappa}$ as the minimum number of iterations to reach the termination criterion $\kappa$. Since the improvement function is defined over the best feasible objective, we define the profit as
\begin{equation} \label{eq:profit}  \small
\begin{aligned}
    \text{profit}(\kappa)= f(x^{+}_{t_{\kappa}})-\kappa \cdot t_{\kappa} \\
    t_{\kappa} =\min t \text{ s.t.\ } \mathcal{\alpha}_t(x_t)<\kappa
\end{aligned}
\end{equation}
where $x^{+}_{t_{\kappa}}= \arg\max_{i\le t_{\kappa} -1}\mu_{t_{\kappa}-1}(x_i)$ and $\kappa$ can be viewed as a cost of evaluating the function set for the optimization routine. A small $\kappa$ implies the computation cost for the objective function is trivial compared with the improvement in the target function. Meanwhile, with a small $\kappa$, the termination will occur after a large number of iterations but also bring an accurate estimation of the target value. Figure~\ref{fig:profitcurve} presents an empirical profit curve for varying values of $\kappa$, showing a consistently higher profit for our proposed acquisition when using a $\kappa$ threshold termination criterion.

\vspace{-0.5cm}
\subsection{Theoretical Properties} \label{sec:theoreticalpart}
In this section, we present the theoretical properties of corrected EI under heteroskedastic noisy outputs, i.e. $y_t=f(x_t)+\varepsilon_t$, which are more natural for real-world applications. The noise term $\varepsilon_t$ is assumed to be Gaussian distributed with a known variance proxy $\upsilon_t^2$. The objective function $f$ is assumed to be smooth according to the reproducing kernel Hilbert space (RKHS) associated with a GP kernel. Similar to \citet{srinivas2010gaussian, wang2014theoretical}, the kernel is assumed to be bounded as $k(x,x)\le 1$. We draw inspiration from \cite{nguyen2017regret} and derived the regret bound in a similar way for our proposed acquisition function. We begin this section with a brief introduction to some important lemmas from existing works. Then we show that, under some mild assumptions, our acquisition function reaches a sublinear convergence rate for the squared exponential kernel similar to the standard EI.

\subsubsection{Auxiliary Definitions and Lemmas}
\begin{Definition}
The maximum information gain after $T$ rounds, namely $\gamma_T$, is defined as:
\begin{equation}  \small
    \gamma_T:=\max _{A \subset D:|A|=T} I\left(\boldsymbol{y}_A ; \boldsymbol{f}_A\right)=\max _{A \subset D:|A|=T} H\left(\boldsymbol{y}_A\right)-H\left(\boldsymbol{y}_A \mid \boldsymbol{f}_A\right)
\end{equation}
where $H\left(\boldsymbol{y}_A\right)$ is the marginal entropy of the observations $\boldsymbol{y}_A=(y_1, \cdots, y_T)$ and $H\left(\boldsymbol{y}_A \mid \boldsymbol{f}_A\right)$ is the conditional entropy of the observations $\boldsymbol{y}_A $ given the corresponding function values $\boldsymbol{f}_A $.
\end{Definition}
\begin{lemma} \label{lemma:1}
[Theorem 6 of \cite{srinivas2010gaussian}]. Let $\delta \in (0,1)$ and assume that the noise variables $\varepsilon_t$ are uniformly bounded by the maximum standard deviation of the observation noise $\upsilon_{\max}=\max(\upsilon_1, \cdots, \upsilon_T)$. Define $\beta_t=2\|f\|_k^2+300 \gamma_t \ln ^3\left(\frac{t}{\delta}\right)$, then
\begin{equation}  \small
p\left(\forall t, \forall x \in \mathcal{X},\left|\mu_{t-1}(x)-f(x)\right| \leq \sqrt{\beta_t} \sigma_{t-1}(x)\right) \geq 1-\delta
\end{equation}
\end{lemma}

\noindent Lemma 7 of \citet{nguyen2017regret} has provided the regret bound for the variance of any arbitrary set of points (not just for selected points $x_t$)  with $\gamma_T$ under the homoskedastic noise setting. In the following lemma, we generalize their findings to accommodate scenarios involving bounded heteroskedastic additive Gaussian observation noise (detailed proof in Appendix \ref{Appendix:lemma}).
\begin{lemma} \label{lemma:4}
The sum of the predictive variances is bounded by the maximum information gain $\gamma_T$. That is for $\forall x \in \mathcal{X}$, it holds that
\begin{equation}  \small
 \sum_{t=1}^T \sigma_{t-1}^2(x)\le \frac{2 }{\log(1+ \upsilon_{\max}^{-2})} \gamma_T 
\end{equation}
where $\upsilon_{\max} = \max(\upsilon_1, \cdots, \upsilon_T)$ is the maximum standard deviation of the additive Gaussian observation noise.
\end{lemma}

\subsubsection{Upper bound for simple regret $r_t$}
Let $x_t$ be the point selected by our acquisition function, then the cumulative regret $R_t$ is the sum of the instantaneous regrets $r_{t}$: $R_t = \sum_{i=1}^{t}r_t$ where $r_{t} = f(x^{\ast})-f(x_t)$. We start the proof sketch by considering breaking down $r_t$ into:
\begin{equation} \small
\begin{aligned}
r_{t} = \underbrace{f(x^{\ast})-f( x_t^+)}_{\textbf{term 1}}-(\underbrace{f(x_t)-f( x_t^+)}_{\textbf{term 2}})\label{eq:simpleRegret}
\end{aligned}
\end{equation}
\begin{lemma} \label{lemma:9}
Let $\kappa >0$ be a pre-defined stopping threshold on the acquisition function $\mathcal{\alpha}_t^{C}(x)$, if $\mu_{t-1}(x_t) \le \mu_{t-1}(x_t^+)$, then we have that 
\begin{equation}  \small
\mu_{t-1}(x_t^+) - \mu_{t-1}(x_t) \le \sqrt{C} \Tilde{\sigma}_{t-1}(x_t) .
\end{equation}
where $C = \log \left[\frac{2}{\pi \kappa^2}\right]$.
\end{lemma}
\begin{proof}
Set $u_{t-1}(x_t) = \mu_{t-1}(x_t)-\mu_{t-1}(x_t^+)$, using our assumption we have $u_{t-1}(x_t) \le 0 $ by the definition of $x_t^+$. Using Equation~\eqref{def:MEI2}, we obtain
\begin{equation}  \small
\begin{aligned}
\mathcal{\alpha}_t^{C}(x_t) &=  \Tilde{\sigma}_{t-1}(x_t) \tau \left( \frac{u_{t-1}(x_t)}{\Tilde{\sigma}_{t-1}(x_t)} \right) \le \Tilde{\sigma}_{t-1}(x_t) \phi \left( \frac{u_{t-1}(x_t)}{\Tilde{\sigma}_{t-1}(x_t)} \right) \\
\end{aligned}
\end{equation}
where the inequality is led by the fact that $\tau(z)\le \phi(z), \forall z<0$. Thus we obtain that 
\begin{equation}  \small
\begin{aligned}
 \frac{u_{t-1}^2(x_t)}{\Tilde{\sigma}^2_{t-1}(x_t)} \le \log \left[ \frac{\Tilde{\sigma}^2_{t-1}(x_t)}{2\pi \kappa^2}\right]  \le \log \left[\frac{2}{\pi \kappa^2}\right] 
\end{aligned}
\end{equation}    
\noindent
where the second inequality is led by the fact that $\Tilde{\sigma}^2_{t-1}(x) \le \sigma^2_{t-1}(x)+\sigma^2_{t-1}(x_t^+) -2 \sigma_{t-1}(xx_t^{+})\le 4$ since the kernel satisfies $k(x,x)\le 1$. Define $C = \log \left[\frac{2}{\pi \kappa^2}\right]\ge 0$, thus we conclude the proof that 
\begin{equation} \small
0\le \mu_{t-1}(x_t^{+}) - \mu_{t-1}(x_t) \le \sqrt{C} \Tilde{\sigma}_{t-1}(x_t).  
\end{equation}
\end{proof}
\begin{lemma}\label{lemma:10}
Let $\kappa >0$ be a pre-defined stopping threshold on the acquisition function $\mathcal{\alpha}_t^{C}(x)$, $z_{t-1}(x)=\frac{\mu_{t-1}(x)-\mu_{t-1}(x_t^+)}{\Tilde{\sigma}_{t-1}(x)}$ and $\tau(z)=z\Phi(z)+\phi(z)$, we have $\tau(-z_{t-1}(x_t))\le 1+\sqrt{C}$ where $C = \log \left[\frac{1}{\pi \kappa^2}\right]$.
\end{lemma}
\begin{proof}
Notice that function $\tau(z)$ has nice properties depending on the sign of $z$: $\tau(z)\le 1+z, \forall z\ge 0$;  $\tau(z)\le \phi(z), \forall z \le 0$. Thus, we consider two possible cases for $z_{t-1}(x_t)$:\\
Case 1: Assume $\mu_{t-1}(x_t) \ge \mu_{t-1}(x_t^+)$ which implies $z_{t-1}(x_t) \ge 0 $, thus 
\begin{equation} \small
    \tau (-z_{t-1}(x_t)) \le \phi(-z_{t-1}(x_t)) \le 1
\end{equation}
Case 2: 
Assume $\mu_{t-1}(x_t) \le \mu_{t-1}(x_t^+)$ which implies $z_{t-1}(x_t) \le 0 $, thus using Lemma \ref{lemma:9}, we have that
\begin{equation} \small
    \tau (-z_{t-1}(x_t)) \le 1 + \frac{\mu_{t-1}(x_t^+)-\mu_{t-1}(x_t)}{\Tilde{\sigma}_{t-1}(x_t)} \le 1+\sqrt{C}
\end{equation}
Clearly, for both cases, we have $\tau(-z_{t-1}(x_t))\le 1+\sqrt{C}$.
\end{proof}

The lemma~\ref{lemma:6} below considers the lower bound for the acquisition function $\mathcal{\alpha}_t^{C}$ under the noisy setting (detailed proof in Appendix \ref{Appendix:lemma}). 
\begin{lemma} \label{lemma:6}
Let $\delta \in (0,1)$. For $x\in \mathcal{X}, t\in \mathcal{N}$, set $I_t^{C}(x)=\max\{0,f(x)-f(x_t^+)\}$, then with probability at least $1-2\delta$ we have
\begin{equation} \small
\mathcal{\alpha}_t^{C}(x) \ge \max\{ I_t^{C}(x) - \sqrt{\beta_t} \left(\sigma_{t-1}(x) + \sigma_{t-1}(x_t^+)\right),0\}.
\end{equation}
\end{lemma}

\noindent Then we consider finding an upper bound for term 1 and term 2 in simple regret given by Equation~\eqref{eq:simpleRegret}. Lemma~\ref{lemma:7} (see Appendix \ref{Appendix:lemma}) provides an upper bound for term 1.
\begin{lemma}  \label{lemma:7}
Let $\delta \in(0,1)$. Then with a probability of at least $1-2\delta$, we have 
\begin{equation}  \small
\begin{aligned}
f(x^{\ast})-f(x_t^+) \le \sqrt{\beta_t} \left(\sigma_{t-1}\left(x^{\ast}\right) + \sigma_{t-1}(x_t^{+})\right)+ \Tilde{\sigma}_{t-1}(x_t) \tau(z_{t-1}(x_t)) .
\end{aligned}
\end{equation}
\end{lemma}

\noindent For term 2, we have that
\begin{equation}  \small
\begin{aligned}
f(x_t^{+})-f(x_t) & = \Tilde{\sigma}_{t-1}(x_t)\left[-z_{t-1}(x_t) \right] \text{   by } z=\tau(z)-\tau(-z) \\
&= \Tilde{\sigma}_{t-1}(x_t) \left( \tau \left(-z_{t-1}(x_t)\right)- \tau\left( z_{t-1}(x_t)\right) \right) .
\end{aligned}
\end{equation}
Finally, by Lemma~\ref{lemma:10} and~\ref{lemma:7}, we obtain 
\begin{equation} \small
\begin{aligned}
 r_t \le f(x^{\ast})-f( x_t^+)+f( x_t^+) - f(x_t) 
   &\le \Tilde{\sigma}_{t-1}(x_t) \tau \left(-z_{t-1}(x_t)\right)+ \sqrt{\beta_t}\left( \sigma_{t-1}(x^{\ast})+ \sigma_{t-1}(x_t^+)\right) \\
  & \le (1+\sqrt{C})\Tilde{\sigma}_{t-1}(x_t)+\sqrt{\beta_t}\left( \sigma_{t-1}(x^{\ast})+ \sigma_{t-1}(x_t^+)\right).
\end{aligned}
\end{equation}
Furthermore, we note that $ \sigma^2_{t-1}(x_t) \sigma^2_{t-1}(x_t^+)-  \sigma^2_{t-1}(x_tx_t^+) \ge 0 $ because of the positive semi-definiteness of the kernel matrix. Thus we have that
\begin{equation} \small
\begin{aligned}
\Tilde{\sigma}_{t-1}(x_t)  = \sqrt{\sigma^2_{t-1}(x_t)+ \sigma^2_{t-1}(x_t^+)-2 \sigma_{t-1}(x_t x_t^+) } 
& \le \sqrt{\sigma^2_{t-1}(x_t)+ \sigma^2_{t-1}(x_t^+) + 2 \sigma_{t-1}(x_t)\sigma_{t-1}(x_t^+)} \\
 & \le \sigma_{t-1}(x_t)+ \sigma_{t-1}(x_t^+) .
\end{aligned}
\end{equation}
Then we can write an upper bound of the simple regret as
\begin{equation} \small
\begin{aligned}
r_t  \le \underbrace{( \sqrt{1} +\sqrt{C}) \sigma_{t-1}(x_t)}_{A_t} + \underbrace{\sqrt{\beta_t}\sigma_{t-1}(x^{\ast})}_{B_t} + \underbrace{(1+ \sqrt{C}+\sqrt{\beta_t})\sigma_{t-1}(x_t^+)}_{C_t} .  \label{ineq:rt}
\end{aligned}
\end{equation}

\subsubsection{Upper bounding the cumulative regret $R_T$}
We now look at the cumulative regret $R_T = \sum_{t=1}^T r_t$
\begin{equation}  \small
\begin{aligned}
& R_T \le \sum_{t=1}^T A_t + \sum_{t=1}^T B_t + \sum_{t=1}^T C_t.   
\end{aligned}
\end{equation}    
Using the Cauchy-Schwartz inequality that $\sum_{i=1}^n a_i^2\le n\left(\sum_{i=1}^n a_i^2\right)$, $\beta_T \ge \beta_t, \forall T\ge t$, and Lemma~\ref{lemma:4}, we can bound $ \sum_{t=1}^T B_t$ with the following
\begin{equation} \small
\begin{aligned}
&\sum_{t=1}^T B_t \le \sqrt{T \sum_{t=1}^T B_t^2} \le \sqrt{\frac{2\beta_T T\gamma_T}{\log(1+\upsilon_{\max}^{-2})}} .
\end{aligned}
\end{equation}
Similarly, we can bound $ \sum_{t=1}^T C_t$ and $ \sum_{t=1}^T A_t$ with the following 
\begin{equation} \small
\begin{aligned}
\sum_{t=1}^T A_t^2  &\le 2(1+C) \sum_{t=1}^T \sigma^2_{t-1}(x_t) \le \frac{4(1+C)\gamma_T}{\log(1+\upsilon_{\max}^{-2})} , \\
\sum_{t=1}^T C^2_t & \le 3(1+C+\beta_T) \sum_{t=1}^T \sigma^2_{t-1}(x_t) \le \frac{6(1+C+\beta_T)\gamma_T}{\log(1+\upsilon_{\max}^{-2})}.
\end{aligned}
\end{equation}
Using the Cauchy-Schwartz inequality again, we get
\begin{equation} \small
\begin{aligned}
 \sum_{t=1}^T A_t  \le \sqrt{T \sum_{t=1}^T A_t^2} \le \sqrt{\frac{4(1+C)T\gamma_T}{\log(1+\upsilon_{\max}^{-2})}} \text{ and } 
\sum_{t=1}^T C_t  \le \sqrt{T \sum_{t=1}^T C_t^2} \le \sqrt{\frac{6(1+C+\beta_T)T\gamma_T}{\log(1+\upsilon_{\max}^{-2})}} . 
\end{aligned}
\end{equation}
Combining the above equations, we obtain our regret bound
\begin{equation} \small
\begin{aligned}
R_T \le \sqrt{\frac{ 2 T\gamma_T}{\log(1+\upsilon_{\max}^{-2})}} \left( \sqrt{\beta_T}+ \sqrt{2(1+C)} + \sqrt{3(1+C+\beta_T)} \right) ,
\end{aligned}
\end{equation}
where $C = \log \left[\frac{2}{\pi \kappa^2}\right]$, $\kappa$ is a pre-defined constant to terminate the optimization, $\beta_T$ is in the form of {\small$\mathcal{O}\left(\left(\log T\right)^3\right)$}, and $\upsilon_{\max} = \max(\upsilon_1, \cdots, \upsilon_T)$ is the maximum standard deviation of the noise. We can see that the regret bound for our proposed acquisition is equivalent to that of standard EI. The bound for $\gamma_T$ relies on the chosen kernel i.e. {\small$\gamma_T \sim \mathcal{O}\left(\left(\log T\right)^{d+1}\right)$} for squared exponential kernel. Therefore, when choosing a squared exponential kernel for the GP model, we achieve a sublinear rate {\small$R_T\sim \mathcal{O} \left( \sqrt{T \left(\log T\right)^{d+4}} \right)$}.

\section{Experiments}
In this section, we present our empirical results using the experiments on the benchmark, synthetic functions, and compression tasks. 

\subsection{Benchmark objective functions}
\begin{figure}[h]\footnotesize
\vspace{-0.3cm}
\begin{minipage}[H]{0.23\linewidth}
\centering
\includegraphics[width=1.5in]{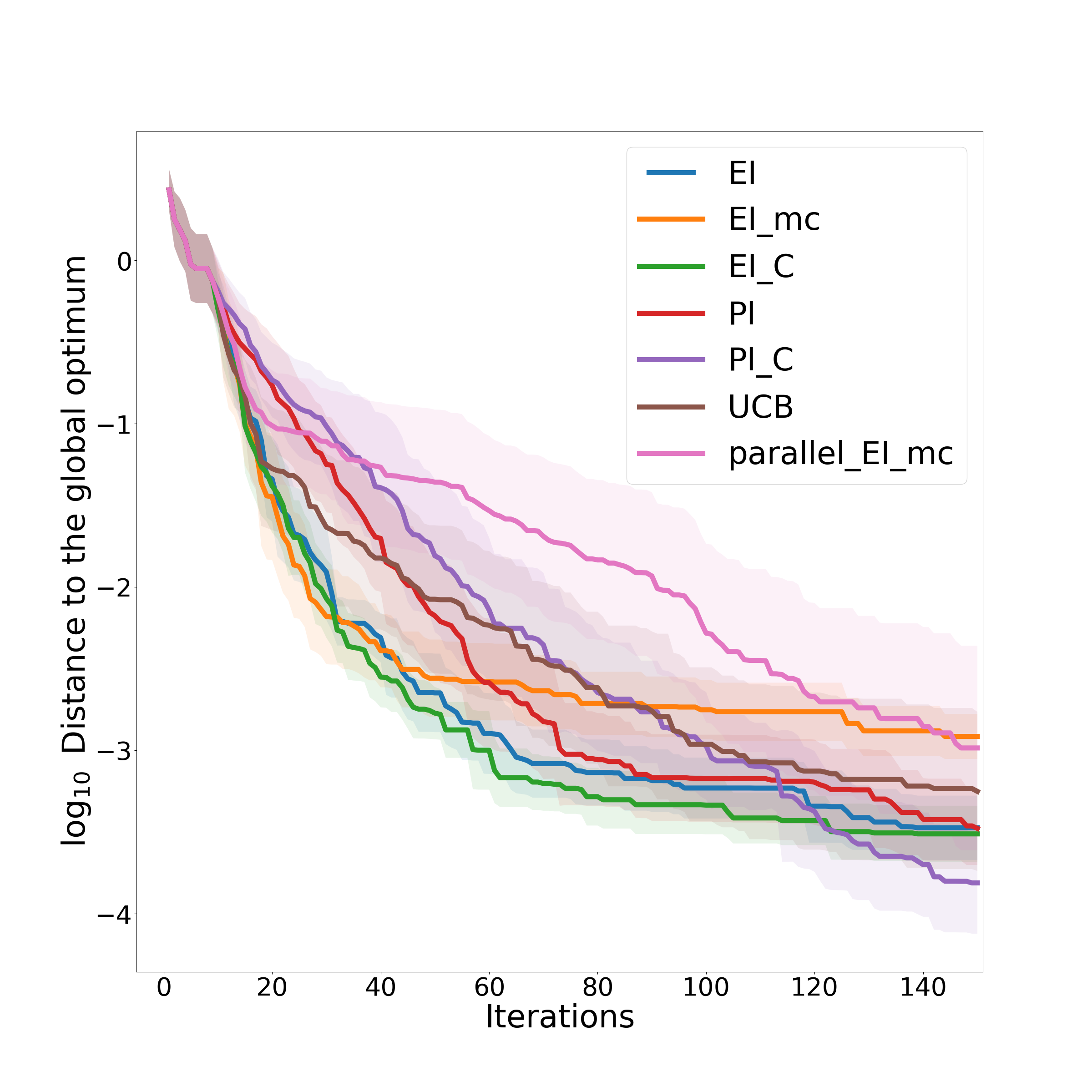}
\caption*{(a) \textit{Hartmann3d}}
\end{minipage}%
\begin{minipage}[H]{0.23\linewidth}
\centering
\includegraphics[width=1.5in]{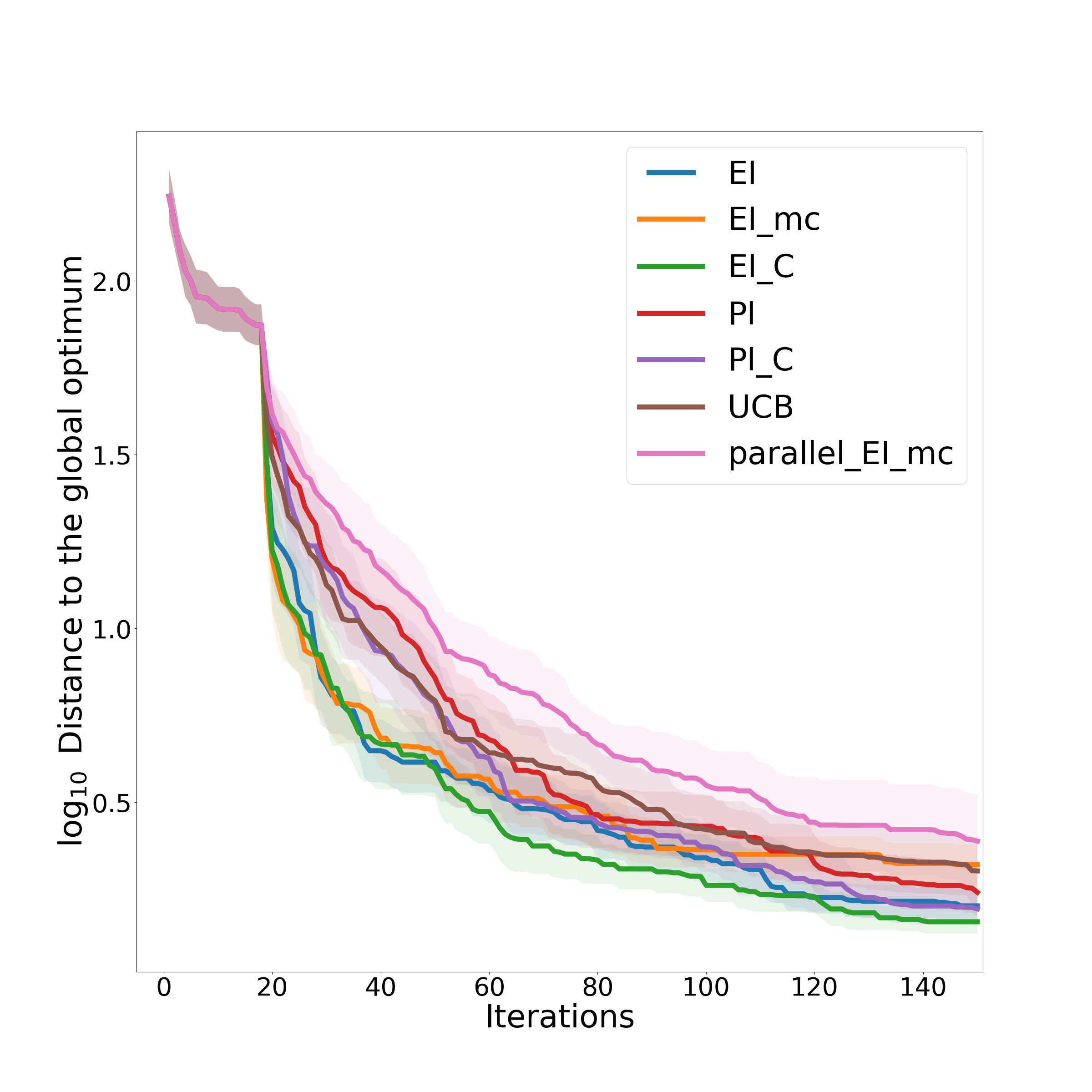}
\caption*{(b) \textit{Griewank}($d=6$)}
\end{minipage}
\begin{minipage}[H]{0.23\linewidth}
\centering
\includegraphics[width=1.5in]{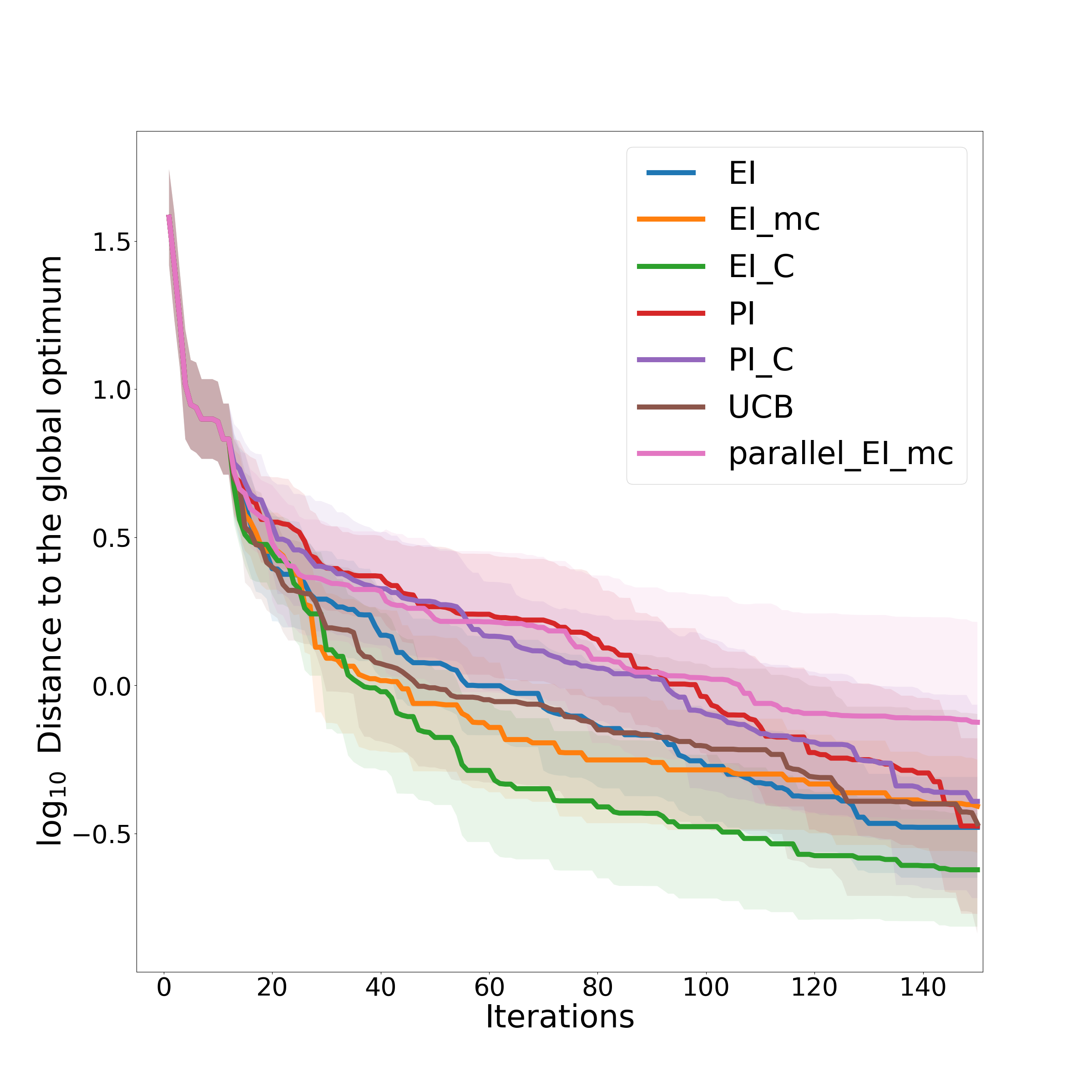}
\caption*{(c) \textit{Levy}($d=4$)}
\end{minipage}
\begin{minipage}[H]{0.23\linewidth}
\centering
\includegraphics[width=1.5in]{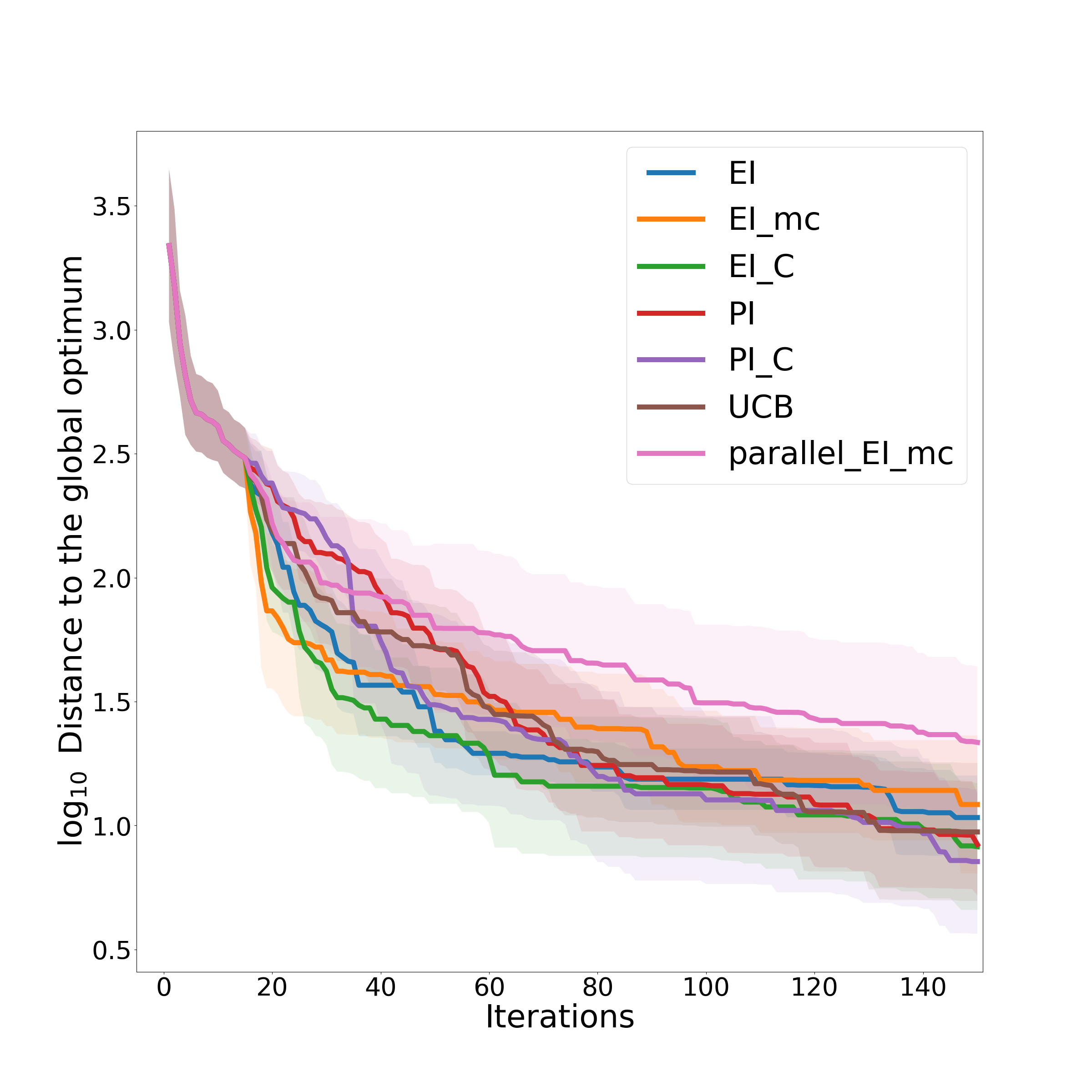}
\caption*{(d) \textit{Powell}($d=5$)}
\end{minipage}
\\
\begin{minipage}[H]{0.23\linewidth}
\centering
\includegraphics[width=1.5in]{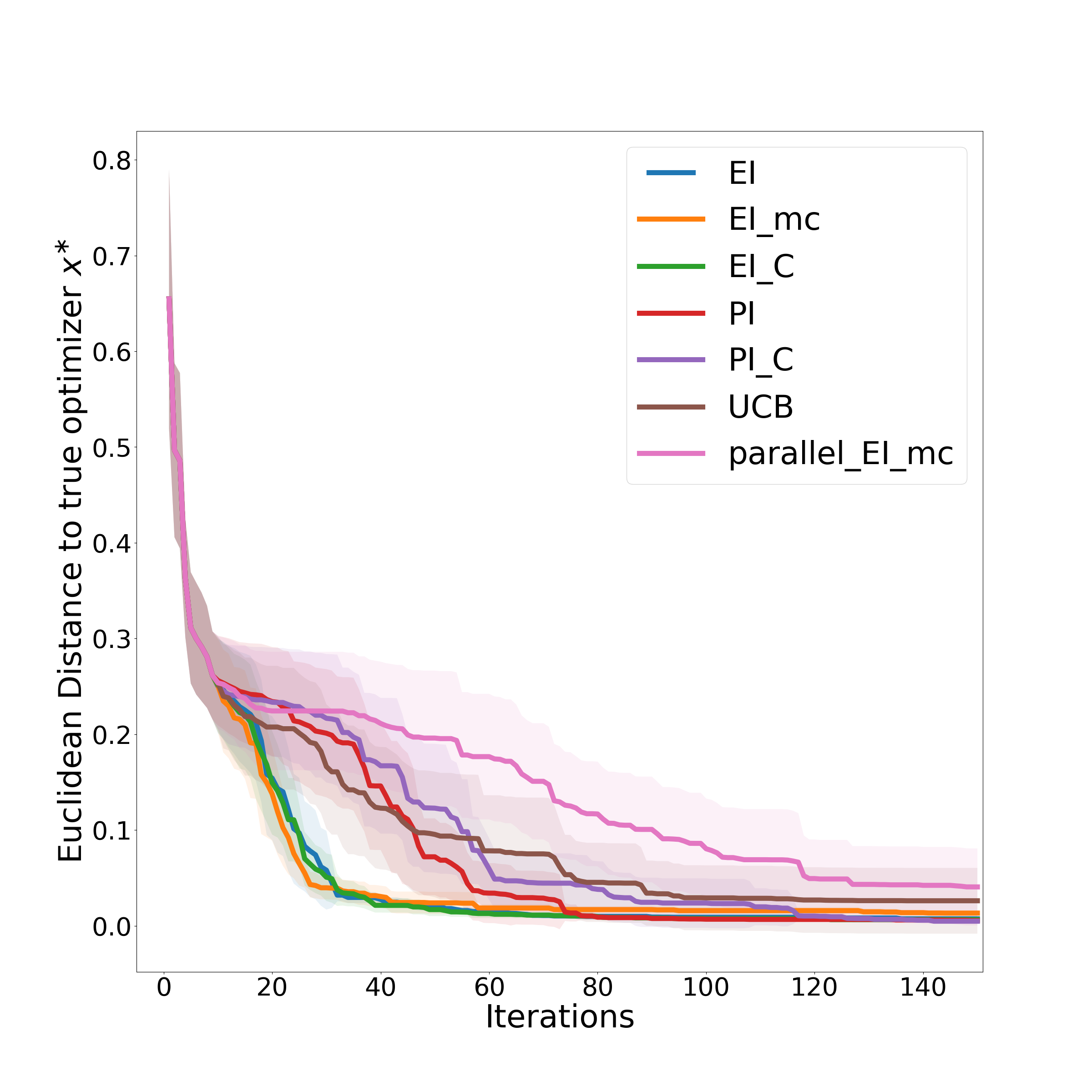}
\caption*{(e) \textit{Hartmann3d}}
\end{minipage}
\begin{minipage}[H]{0.23\linewidth}
\centering
\includegraphics[width=1.5in]{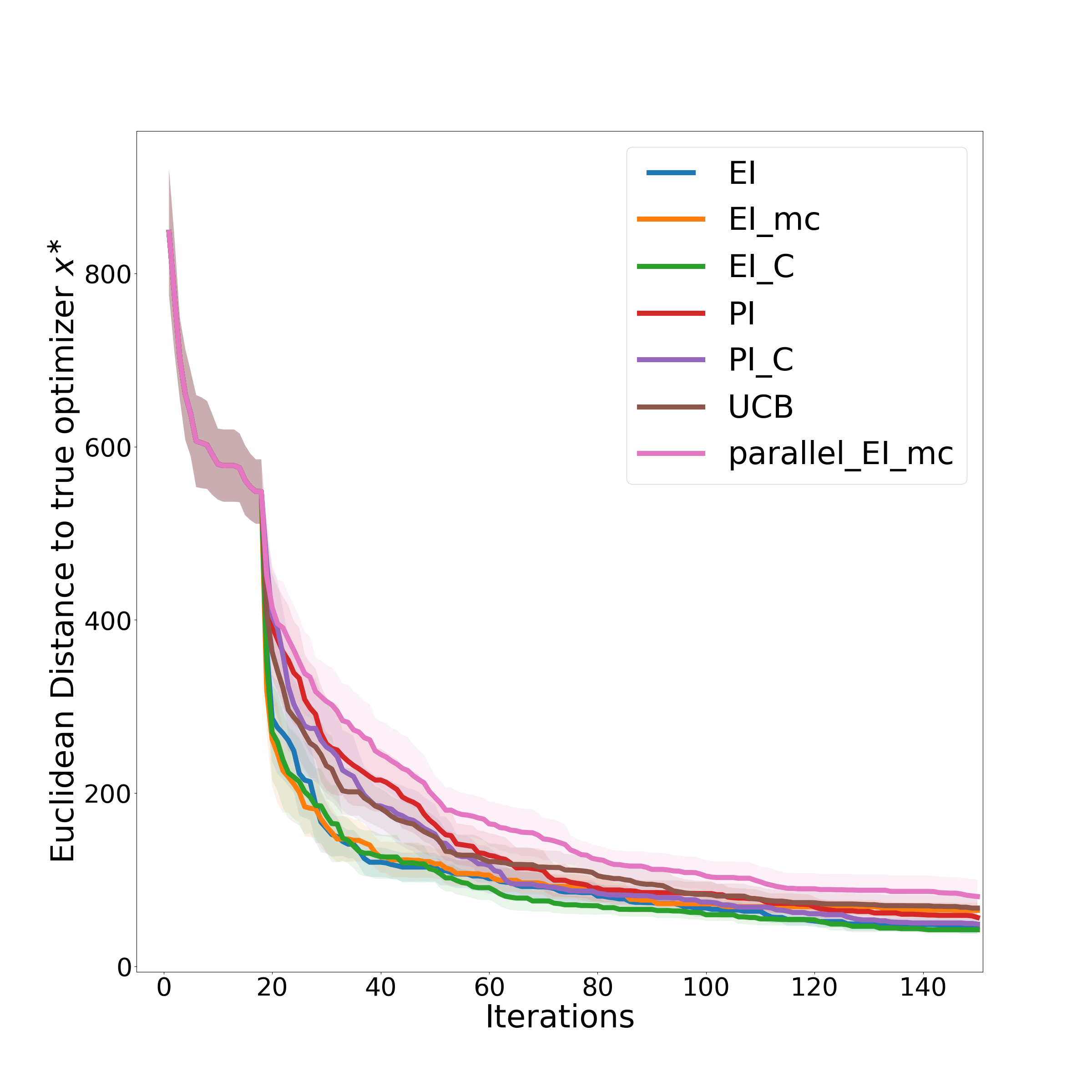}
\caption*{(f) \textit{Griewank}($d=6$)}
\end{minipage}
\begin{minipage}[H]{0.23\linewidth}
\centering
\includegraphics[width=1.5in]{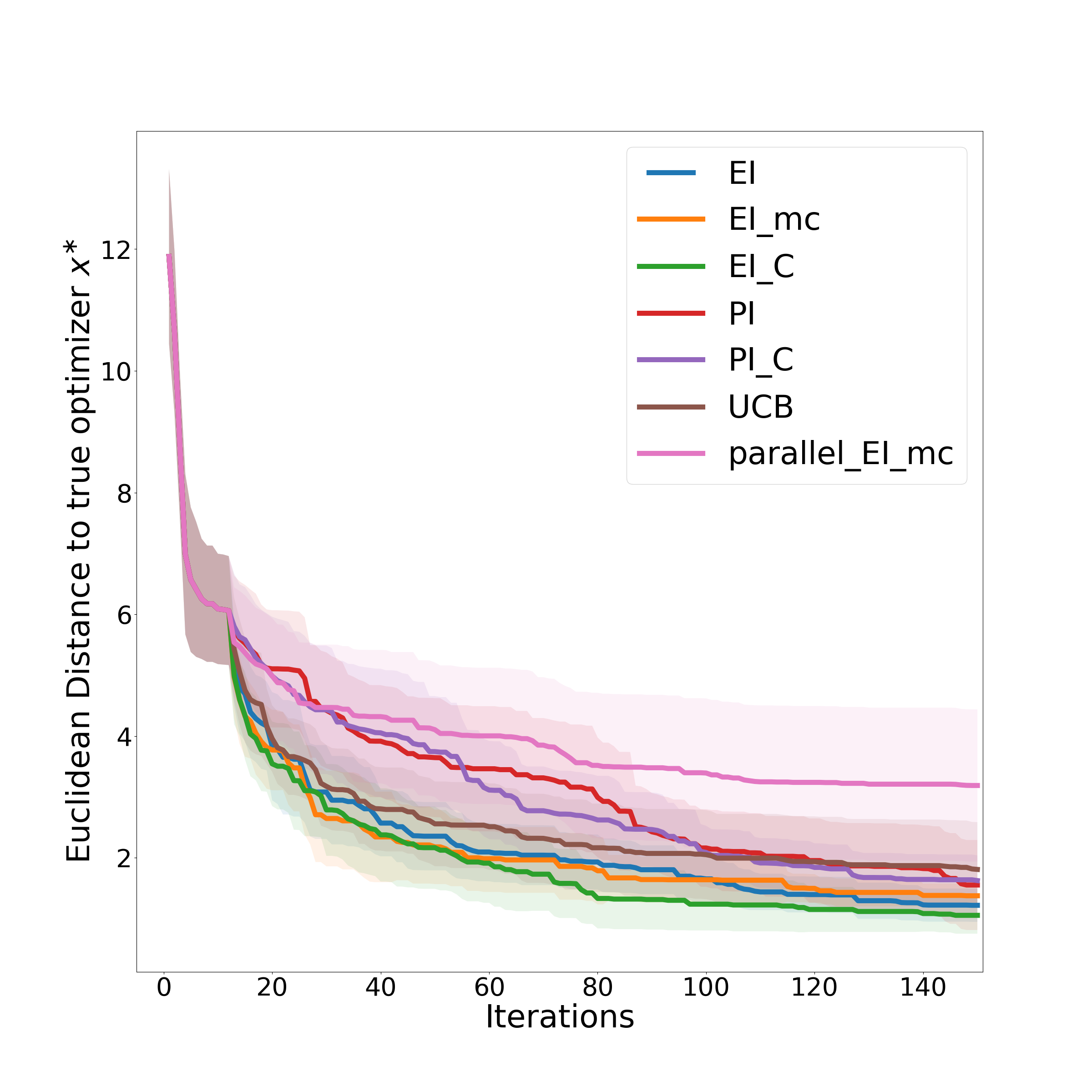}
\caption*{(g) \textit{Levy}($d=4$)}
\end{minipage}
\begin{minipage}[H]{0.23\linewidth}
\centering
\includegraphics[width=1.5in]{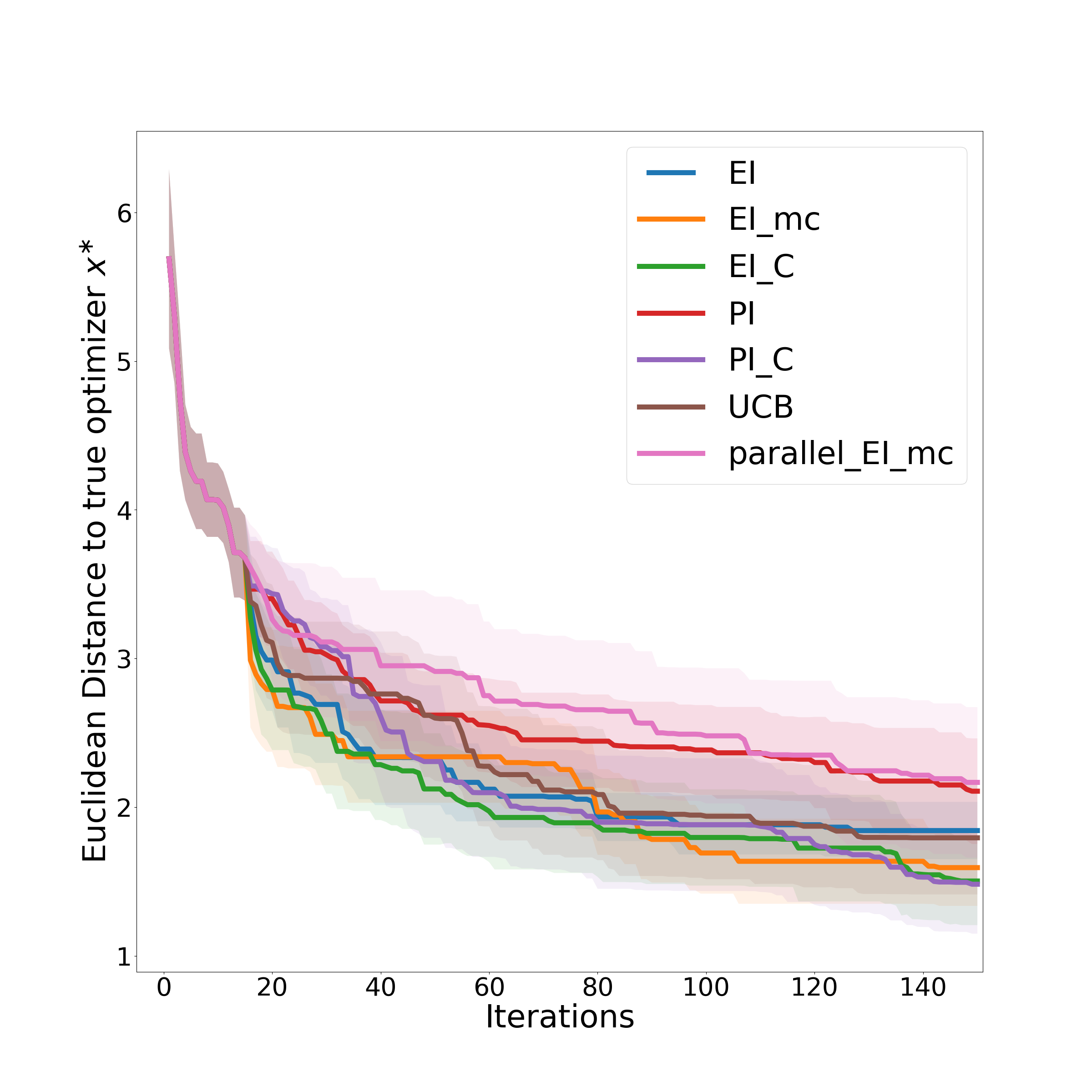}
\caption*{(h) \textit{Powell}($d=5$)}
\end{minipage}
\caption{Comparison of methods for Benchmark objective functions. Figures (a)-(d) show how the mean and 95\% confidence bound (shaded region) of the distance between the best feasible objective and the global optimum changes with each iteration of optimization. Figures (e)-(f) visualize the variation of the $L_2$ distance between the best point and the global optimizer $x^{\ast}$.}
\label{fig:synthetic}
\end{figure} 
\vspace{-0.1cm}

In this section, we compare our acquisition method with other methods including Corrected-PI~\citep{ma2019bayesian}, UCB, EI, PI, MC-based noisy EI~\citep{letham2019constrained}, and parallel MC-based noisy EI~\citep{balandat2020botorch} on several benchmark objective functions including \textit{Hartmann3d}, \textit{Griewank}($d=6$), \textit{Levy}($d=4$), and \textit{Powell}($d=5$). Our objective is to identify the optimizer that minimizes the values of these functions. We use the framework of \textit{BoTorch}\footnote{\textit{BoTorch}~\citep{balandat2020botorch} is a state-of-the-art open-source Bayesian optimization software package with support for various acquisition functions.} for implementing those acquisition functions. Two evaluation metrics are considered: the log distance to the global optimum $\log_{10}\left(f\left(x^{\ast}\right)-f\left(x^{+}_t\right)\right)$ and the $L_2$ distance to the global optimizer $x^{\ast}$. We employed the Matérn kernel for our GP model and set the total number of iterations $(T)$ to 150. The length scale parameter of the kernel is optimized by maximum likelihood. To introduce noise in our experiments, each observation noise $\varepsilon_t$ was sampled from a Gaussian distribution with mean 0 and standard deviation $\upsilon_t$ less than or equal to 10\% of the range of the objective function $\left(\max f\left(x\right) - \min f\left(x\right)\right)$. Before fitting the model, the outputs along with observation noises are standardized, and the inputs are normalized to $[0,1]^d$ based on the minimum and maximum values. All experiments were repeated 15 times for each benchmark function with a quasirandom sequence of size $3d$ as the initialization for our GP model. 

Results in Figure~\ref{fig:synthetic} indicate that the corrected EI outperforms EI for those benchmark objective functions under our noisy settings. For \textit{Powell} function, the corrected PI slightly outperforms our proposed method. Overall, our proposed method demonstrates good performance on those benchmark functions. Furthermore, we assess the computational cost associated with these acquisition functions. Typically, this cost involves inferring the hyperparameters of the GP model, and maximizing the acquisition function in order to propose a candidate point. With analytic acquisition functions, the objective function can be approximated using just one GP model, enabling direct computation of the acquisition value from this model. Consequently, working with this kind of acquisition function is considerably inexpensive. In contrast, MC-type methods are more computationally expensive. For example, MC-based noisy EI relies on multiple noiseless GP models and performs integration by averaging the expected improvement across these models, resulting in higher computational costs. Moreover, parallel MC-based noisy EI assumes that the incumbent best is unknown and uses samples from the joint posterior over the $q$ test points and previously observed points. The integration is computed by averaging the improvements on those samples. In our experiment, we set the number of noiseless models to be $20$ (by default) for MC-based noisy EI, and a quasirandom sequence of size $q=256$ for parallel MC-based noisy EI, resulting in significantly higher computational costs for these two methods compared to analytic acquisition functions.
\subsection{Model Compression}

Various compression techniques have been proposed for DNN models, leading to a smaller model that can be deployed on edge devices with limited memory and computational resources. Low-rank factorization techniques like Singular Value Decomposition (SVD) and Tensor Decomposition can be utilized to construct the compressed network with a low-rank approximation of the original weight matrices. Their rank parameters denoted as $\theta$ can be selected via the BO procedure in order to find a balance between the size and performance of the compressed networks. A scaling scheme proposed by~\cite{ma2019bayesian} is applied to transform the rank parameter from discrete space to continuous domain $[0,1]^{d}$ where $d$ represents the number of dimensions in the parameterization. Let us define $f^{\ast}$ as the original model and $\hat{f}_{\theta}$ as the compressed model, then the objective function with respect to $\theta$ can be specified as
\begin{equation} \small
\gamma \mathcal{L}(\hat{f}_{\theta})+ \mathcal{R}(\hat{f}_{\theta},f^{\ast})
\end{equation}
where $\mathcal{R}(\hat{f}_{\theta},f^{\ast})$ is the compression ratio which is calculated by dividing the size of the compressed network by the size of the original network, and $\mathcal{L}(\hat{f}_{\theta})$ is the error rate of the compressed network. A smaller model with good generalization performance is thus preferred in order to minimize this objective function and $\gamma$ is the trade-off parameter.

\begin{figure*}[h]\small 
\begin{minipage}[h]{0.32\linewidth}
\centering
\includegraphics[width=1.4in]{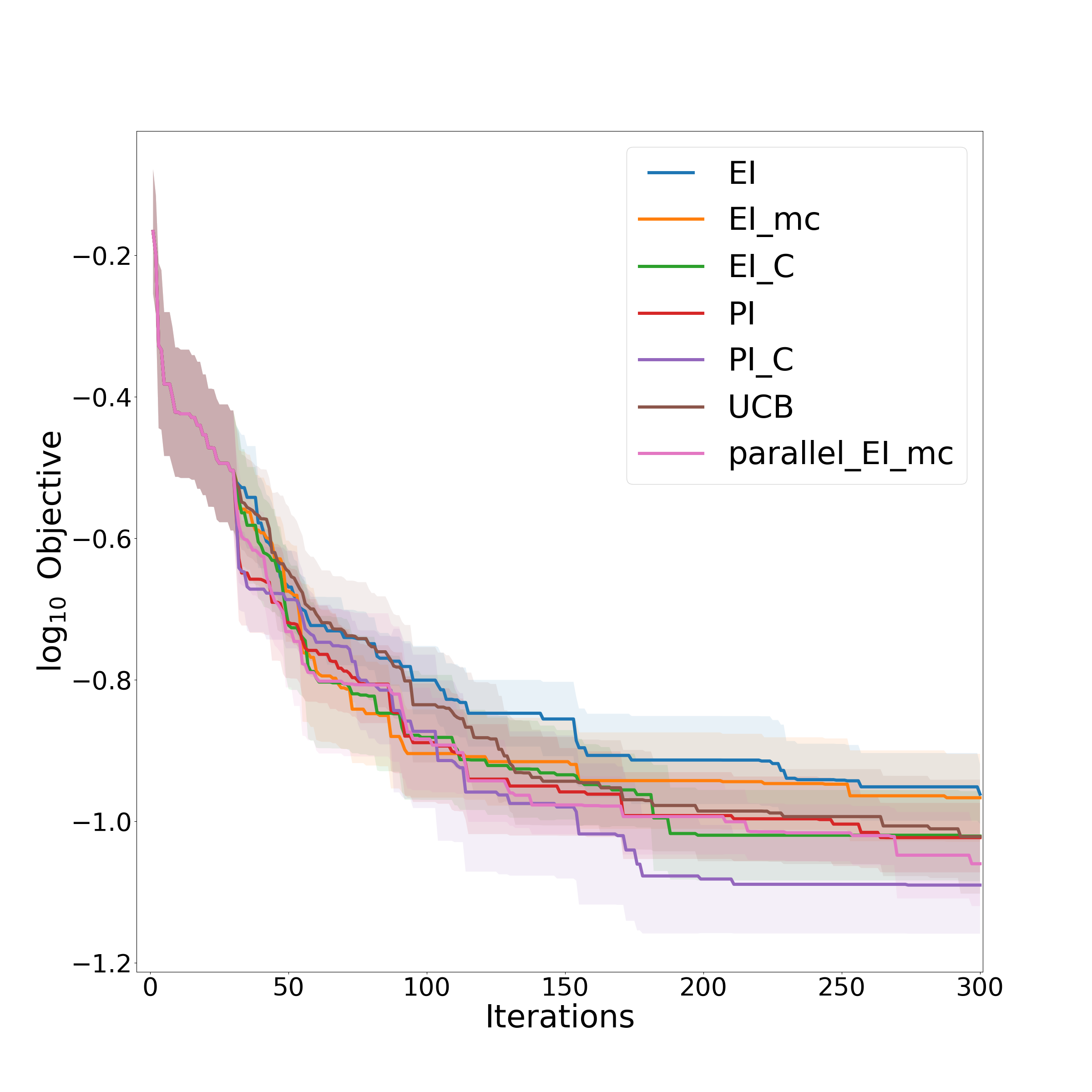}
\caption*{(a) FC3: all iterations }
\end{minipage}%
\begin{minipage}[h]{0.32\linewidth}
\centering
\includegraphics[width=1.4in]{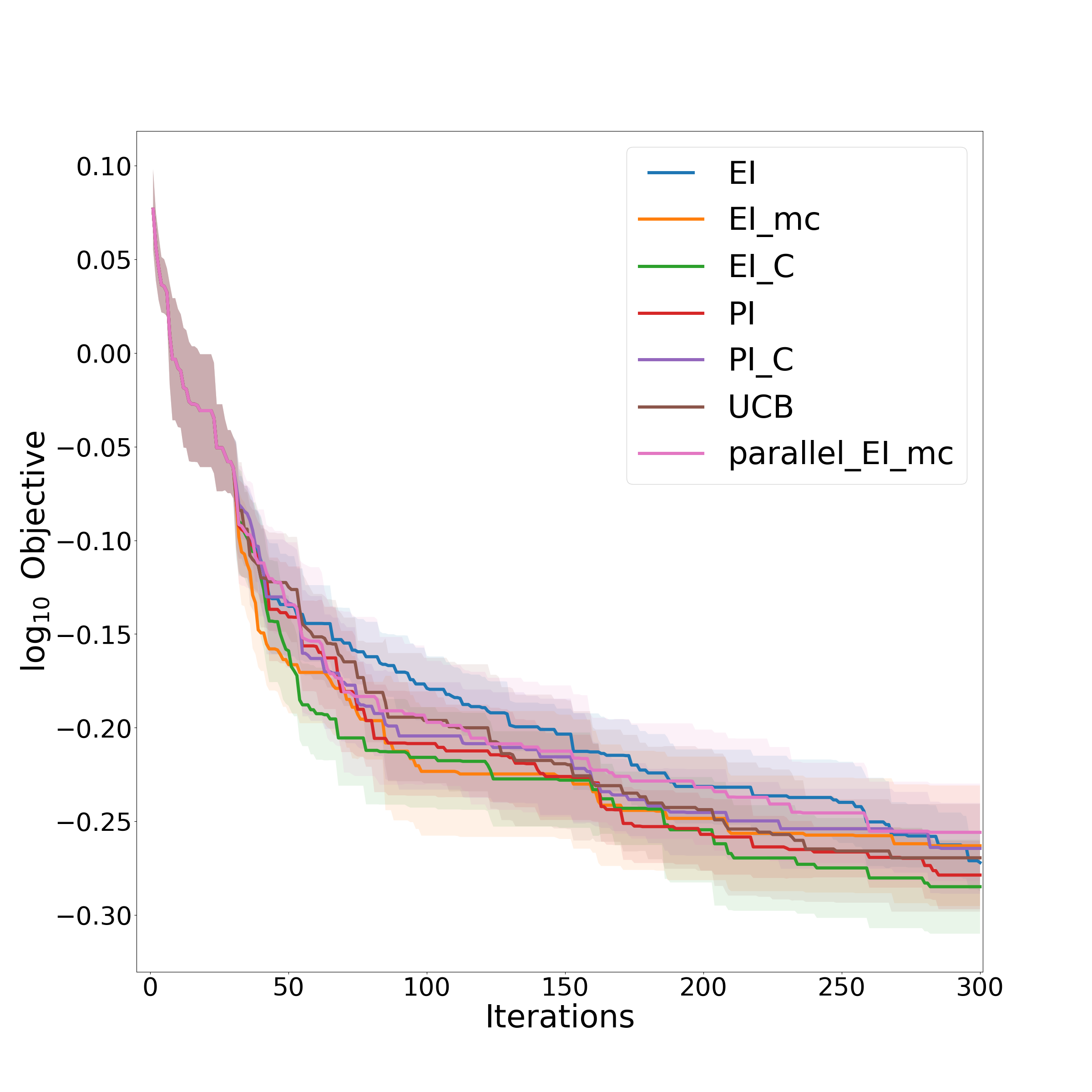}
\caption*{(b) ResNet50: all iterations }
\end{minipage}
\begin{minipage}[h]{0.32\linewidth}
\centering
\includegraphics[width=1.4in]{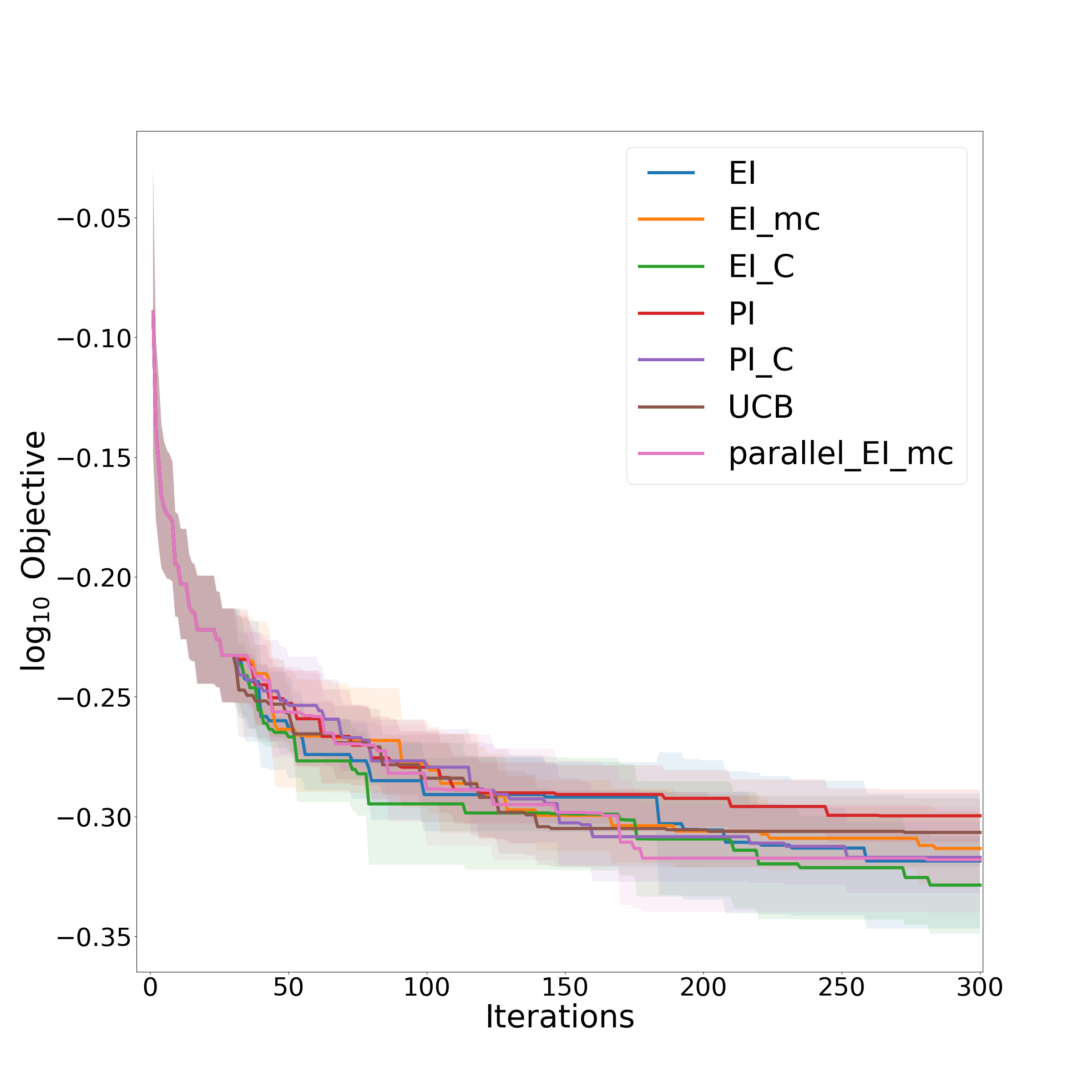}
\caption*{(c) VGG-16: all iterations}
\end{minipage}
\\
\begin{minipage}[h]{0.32\linewidth}
\centering
\includegraphics[width=1.4in]{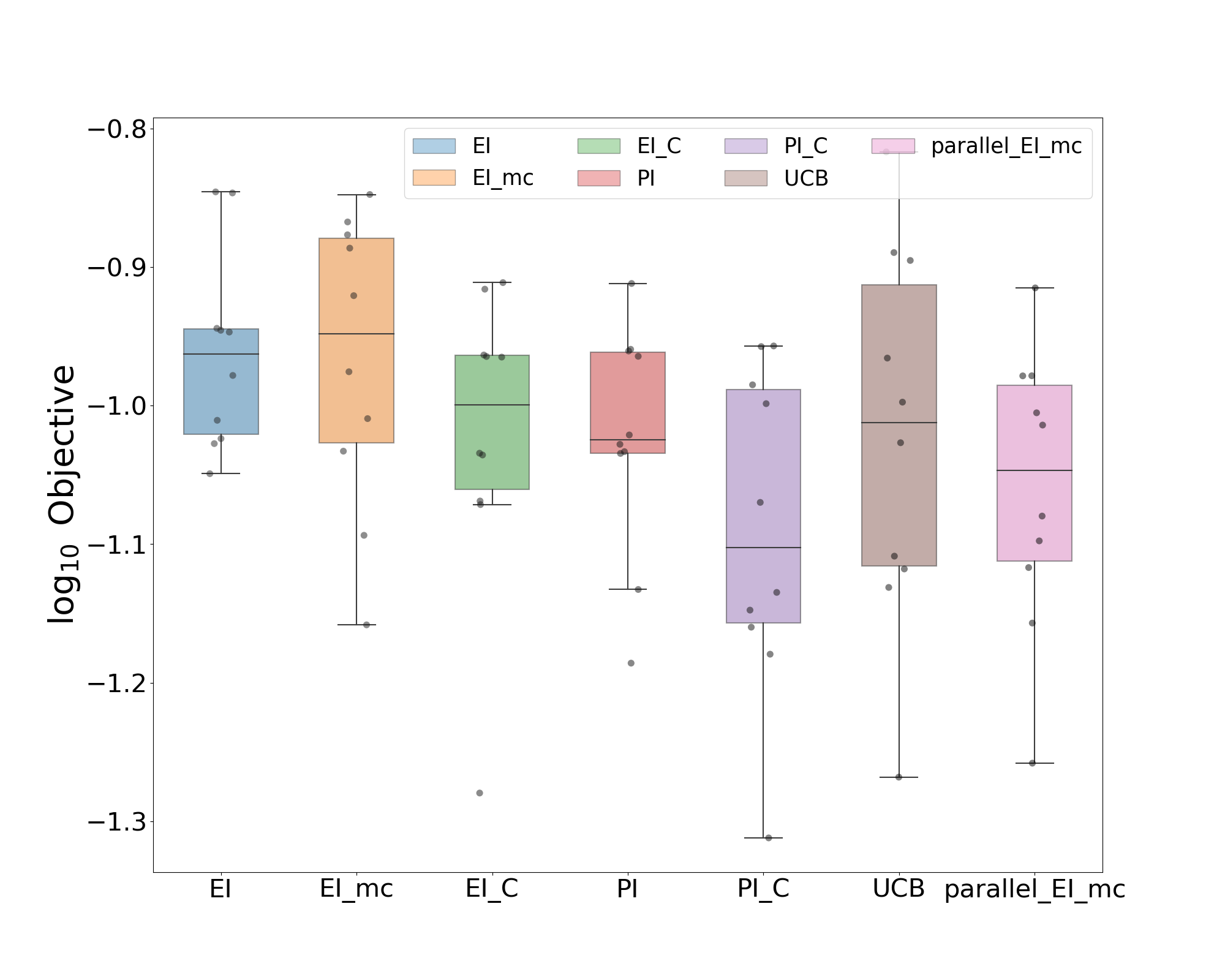}
\caption*{(d) FC3: best of all iterations}
\end{minipage}%
\begin{minipage}[h]{0.32\linewidth}
\centering
\includegraphics[width=1.4in]{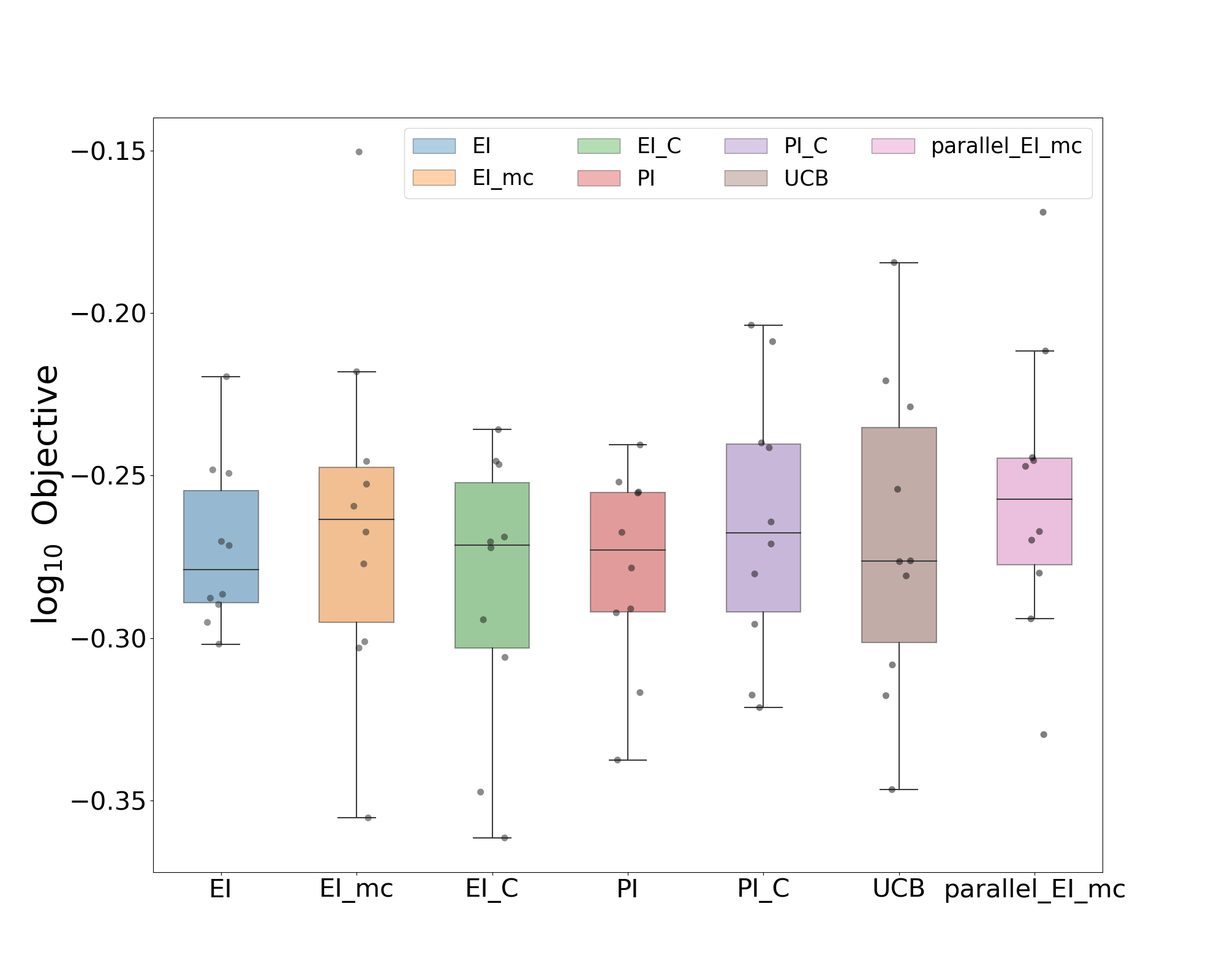}
\caption*{(e) ResNet50: best of all iterations}
\end{minipage}
\begin{minipage}[h]{0.32\linewidth}
\centering
\includegraphics[width=1.4in]{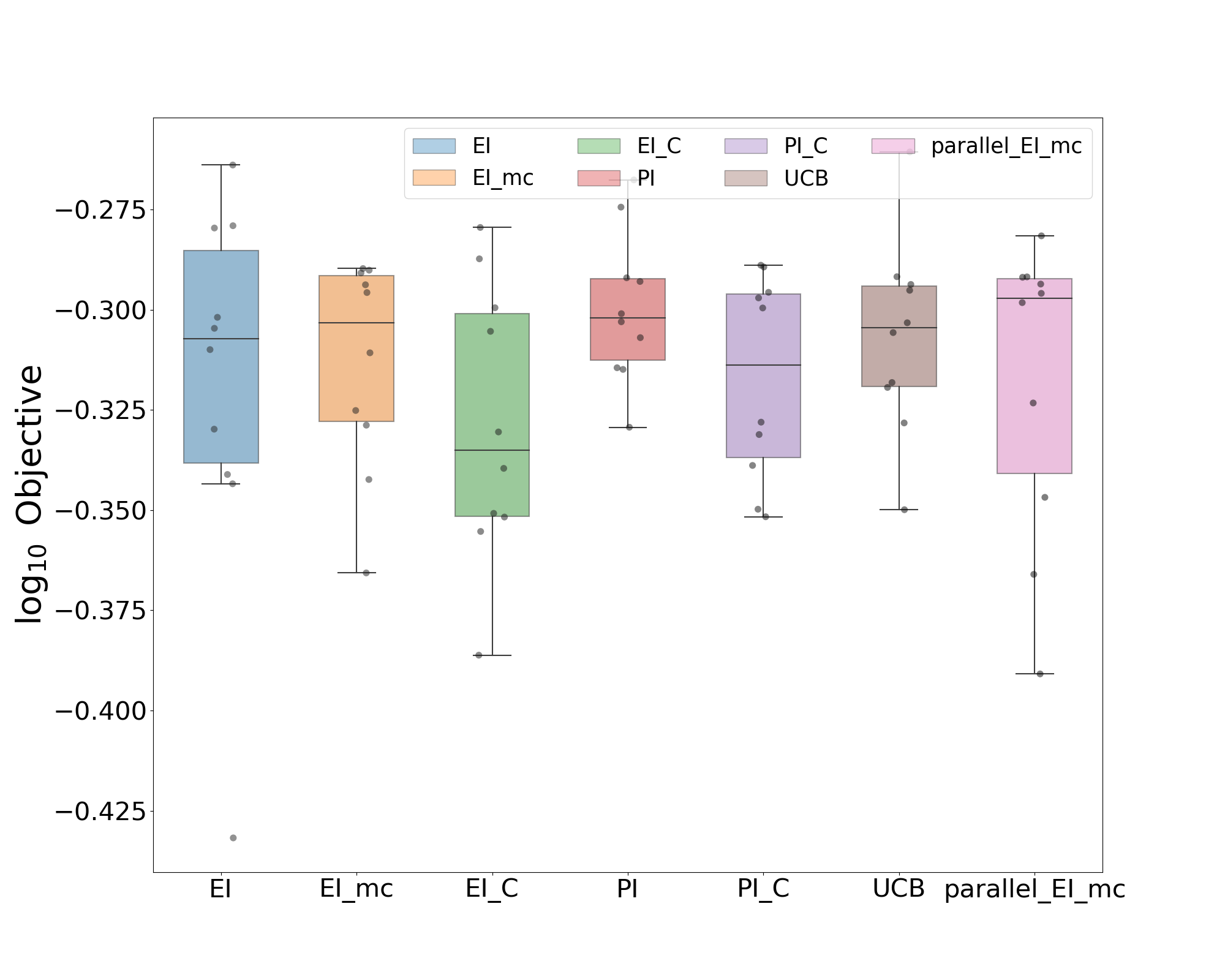}
\caption*{(f) VGG-16: best of all iterations}
\end{minipage}
\caption{Comparison results on the compression tasks of the pre-trained FC3, ResNet50, and VGG-16 model using BO with different acquisition functions. The results are obtained from 10 independent runs. The plots in the first rows show how the mean of the best observation changes over the iterations of BO together with its 95\% confidence interval. The second row shows the best results among all iterations.}
\label{fig:compression}
\end{figure*}
\vspace{-0.2cm}
In our experiments, we compare our approach with other acquisition functions on the compression tasks of several representative neural networks including a 3-layer fully connected network (FC3,~\cite{ma2019bayesian}), ResNet50~\citep{he2016deep}, and VGG-16~\citep{simonyan2014very}. We compress all the layers of FC3 using SVD as proposed in~\cite{denton2014exploiting}. While for the latter two models, we only compress the weights of their convolution layers using tensor decomposition~\citep{tai2016convolutional}. Thus there are 3 compression parameters in the compression task of FC3, 13 parameters in that of VGG-16, and 16 parameters for ResNet50. The FC3 model is pre-trained on the MNIST dataset~\citep{deng2012mnist} while the latter two models are pre-trained on the ImageNet dataset~\citep{russakovsky2015imagenet}. The trade-off parameter $\gamma$ is set to be 1. During each iteration of optimization, we consistently measure the top-1 error rate on $n_t$ randomly selected samples ($20\le n_t \le 50$), resulting in a noisy evaluation of the objective. For each observation, its noise variance $\upsilon_t^2$ is inversely proportional to the sample size $n_t$. For FC3, those samples are randomly drawn from the 10,000 testing images of the MNIST dataset. For the latter two models, the samples are selected from 50,000 validation images of the ILSVRC2012 dataset. We continue using the previous settings for the GP model and the acquisition functions. We run all algorithms for 300 iterations with the first 30 iterations being random initialization. The compression results in Figure~\ref{fig:compression} show our proposed method gives better performance than standard EI under the noisy setting.
\vspace{-0.1cm}
\section{Discussion and Conclusions}
In this paper, we propose a novel acquisition function that addresses the limitation of the analytic EI-type methods in the presence of noise. We correct the closed-form expression of EI to account for the uncertainty introduced by the incumbent best \textemdash a result that has not previously been published despite EI being nearly half a century old \citep{Mockus1978} and one of the most popular acquisition functions even in the presence of observation noise. Additionally, we show that this modified EI retains a convergence rate similar to that of the standard EI, and results in a profit-maximizing convergence criterion under a linear cost model. Our empirical experiments provide evidence that this approach is effective and competitive compared with some popular acquisition functions when dealing with noisy observations. Although there may be cases where our method underperforms corrected-PI or MC-based methods, we believe that it offers valuable insights into the behavior of EI-type methods under noisy observations. One notable contribution of our work is that it fills a gap in the analytic EI-type approaches by directly incorporating the uncertainty of the incumbent best without relying on MC integration when formulating the closed-form expression. This enhancement improves both the efficiency and performance of BO by leveraging the available covariance information from the GP model.

We also notice that the differences in performance among those acquisitions might be negligible for functions in high dimensions~\citep{ma2019bayesian} or extremely noisy observations~\citep{garnettbayesoptbook2022}. In more complex settings of real-world problems, some details about the data information such as the scale of the observation noise are still unknown to us thus it can be challenging to intuit suitable parameters i.e.\ prior distribution for the heteroscedastic GP model without a great deal of knowledge about the data. In the worst case with increasing levels of noise, the GP inference reflects more uncertainty in the objective function regardless of the choice of the acquisition functions. Our work also reflects the importance of combining the covariance information provided by this model in Bayesian optimization. Therefore, we expect more advanced techniques for characterizing and capturing the nature of observation noise, which can be instrumental in constructing a more precise GP model.

\vspace{-0.1cm}
\acks{This work is partially funded by the Research Foundation - Flanders (FWO)
projects G0A1319N and G0G2921N. HZ is supported by the China Scholarship Council.}
\vspace{-0.1cm}
\bibliography{acml23}
\pagebreak

\appendix
\section{Supplements for Lemmas} \label{Appendix:lemma}

\begin{mythm}{2} [Generalized version of Lemma 7 of \cite{nguyen2017regret}] 
The sum of the predictive variances is bounded by the maximum information gain $\gamma_T$. That is for $\forall x \in \mathcal{X}$, it holds that
\begin{equation}
 \sum_{t=1}^T \sigma_{t-1}^2(x)\le \frac{2 }{\log(1+ \upsilon_{\max}^{-2})} \gamma_T 
\end{equation}
where $\upsilon_{\max} = \max(\upsilon_1, \cdots, \upsilon_T)$ is the maximum standard deviation of the additive Gaussian observation noise.
\end{mythm}
\begin{proof}
Let us define $G(x) = \frac{x}{\log(1+x)}$, we notice that $G(x)$ is monotonically increasing when $x\ge 0$ with its minimum value $G(0)=0$. Utilizing this property, we have that $\frac{s}{\log(1+s)} \le \frac{\upsilon_t^{-2}}{\log(1+ \upsilon_t^{-2}) }$ for $s \in [0, \upsilon_t^{-2}]$ and  $s=\upsilon_t^{-2} \sigma_{t-1}^2(x) \le \upsilon_t^{-2} $ since $\sigma_{t-1}^2(x) \le k(x,x) \le 1 $. Define
$\upsilon_{\max} = \max(\upsilon_1, \cdots, \upsilon_T)$ as the maximum standard deviation of the additive Gaussian observation noise, then we can derive that
\begin{equation}
\begin{aligned}
\forall x \in \mathcal{X}, \sum_{t=1}^T \sigma_{t-1}^2(x) & = \sum_{t=1}^T \upsilon_t^{2} \underbrace{\upsilon_t^{-2}  \sigma_{t-1}^2(x) }_{s} \le \sum_{t=1}^T  \upsilon_t^{2} \left(  \frac{\upsilon_t^{-2} \log(1+s )}{\log(1+ \upsilon_t^{-2}) }    \right) \\
& = \sum_{t=1}^T \frac{ \log(1+ \upsilon_t^{-2} \sigma_{t-1}^2 (x) )}{\log(1+ \upsilon_t^{-2}) } \le \frac{2 }{\log(1+ \upsilon_{\max}^{-2})} \frac{1}{2} \sum_{t=1}^T  \log(1+ \upsilon_t^{-2} \sigma_{t-1}^2 (x) ) \\
& \le \frac{2 }{\log(1+ \upsilon_{\max}^{-2})} \gamma_T 
\end{aligned}
\end{equation}
where the last inequality is led by the definition of $\gamma_T$.
\end{proof}

\begin{mythm}{5} 
Let $\delta \in (0,1)$. For $x\in \mathcal{X}, t\in \mathcal{N}$, set $I_t^{C}(x)=\max\{0,f(x)-f(x_t^+)\}$, then with probability at least $1-2\delta$ we have
\begin{equation}
\mathcal{\alpha}_t^{C}(x) \ge \max\{ I_t^{C}(x) - \sqrt{\beta_t} \left(\sigma_{t-1}(x) + \sigma_{t-1}(x_t^+)\right),0\}.
\end{equation}
\end{mythm}
\begin{proof}
If $\Tilde{\sigma}_{t-1}(x)=0$, then we have $\mathcal{\alpha}_t^{C}(x) = I_t^{C}(x)  = 0 $. We now assume $\Tilde{\sigma}_{t-1}(x)>0$ . Set $q = \frac{f(x)-f(x_t^+)}{\Tilde{\sigma}_{t-1}(x)}$ and $\Tilde{z}=\frac{\mu_{t-1}(x)-\mu_{t-1}(x_t^+)}{\Tilde{\sigma}_{t-1}(x)}$, then we have $\Tilde{z}-q=\frac{f(x_t^+)-\mu_{t-1}(x_t^+)-(f(x)-\mu_{t-1}(x))}{\Tilde{\sigma}_{t-1}(x)}$. By Lemma \ref{lemma:1}, we have that $|\Tilde{z}-q| \le \frac{\sigma_{t-1}(x) + \sigma_{t-1}(x_t^+)}{\Tilde{\sigma}_{t-1}(x)}\sqrt{\beta_t}$ holds with probability $1-2\delta$. Denote $m(x) =\frac{\sigma_{t-1}(x) + \sigma_{t-1}(x_t^+)}{\Tilde{\sigma}_{t-1}(x)}$, thus $q-m(x)\sqrt{\beta_t} \le \Tilde{z}$.
If $I_t^{C}(x) = 0$, then the lower bound is trivial as $\mathcal{\alpha}_t^{C}(x)$ is non-negative. Thus suppose $I_t^{C}(x) > 0$. Set $\tau(z)=z\Phi(z)+\phi(z)$, since $\tau(z)$ is non-decreasing for all $z$, we have that 
\begin{equation}
\begin{aligned}
\mathcal{\alpha}_t^{C}(x) \ge {\Tilde{\sigma}_{t-1}(x)} \tau (q-m(x)\sqrt{\beta_t}) & \ge {\Tilde{\sigma}_{t-1}(x)}(q-m(x)\sqrt{\beta_t}) \quad  \text{by } \tau(z) \ge z  \\
& =  I_t^{C}(x) - \sqrt{\beta_t} \left(\sigma_{t-1}(x) + \sigma_{t-1}(x_t^+)\right).
\end{aligned}
\end{equation}
\end{proof}

\begin{mythm}{6} 
Let $\delta \in(0,1)$. Then with a probability of at least $1-2\delta$, we have 
\begin{equation}  \small
\begin{aligned}
f(x^{\ast})-f(x_t^+) \le \sqrt{\beta_t} \left(\sigma_{t-1}\left(x^{\ast}\right) + \sigma_{t-1}(x_t^{+})\right)+ \Tilde{\sigma}_{t-1}(x_t) \tau(z_{t-1}(x_t)) .
\end{aligned}
\end{equation}
\end{mythm}

\begin{proof}
By Lemma~\ref{lemma:6} and $I_t^M(x)=\max\{0, f(x)-f(x_t^+)\}$, we have that 
\begin{equation}  \small
\begin{aligned}
 f(x^{\ast})-f(x_t^+) \le I_t^M(x^{\ast}) \le \sqrt{\beta_t} \left(\sigma_{t-1}(x^{\ast}) + \sigma_{t-1}(x_t^{+})\right)+\mathcal{\alpha}_t^{C}(x^{\ast})
\end{aligned}
\end{equation}
where the second inequality is provided by Lemma \ref{lemma:6}. By the definition of $x_t=\arg\max_{x\in\mathcal{X}}\mathcal{\alpha}_t^M(x)$, we obtain
\begin{equation}  \small
\begin{aligned}
\mathcal{\alpha}_t^{C}(x^{\ast}) \le \mathcal{\alpha}_t^{C}(x_t) = \Tilde{\sigma}_{t-1}(x_t) \tau(z_{t-1}(x_t))
\end{aligned}
\end{equation}    
Thus, we derive the following result by combining the above two inequalities
\begin{equation}  \small
\begin{aligned}
f(x^{\ast})-f(x_t^+) \le \sqrt{\beta_t} \left(\sigma_{t-1}(x^{\ast}) + \sigma_{t-1}(x_t^{+})\right)+ \Tilde{\sigma}_{t-1}(x_t) \tau(z_{t-1}(x_t)) .
\end{aligned}
\end{equation}
This final inequality holds with probability $1-2\delta$.
\end{proof}

\section{Additional simulation results: Functions sampled from Gaussian kernel}
\begin{figure}[H]
\centering
\includegraphics[width=4.5 in]{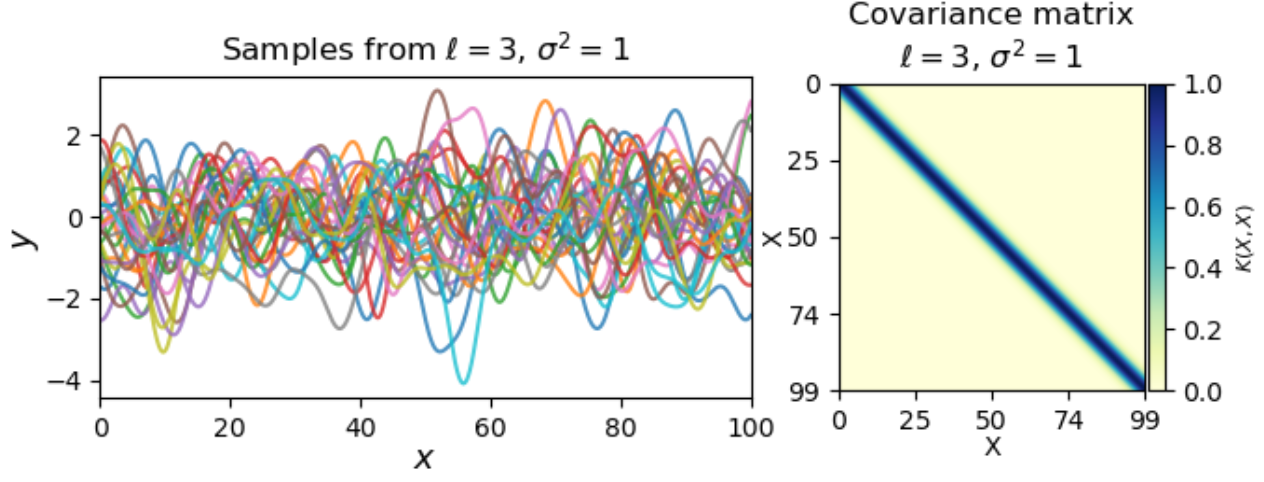}
\caption{ (Left) Samples from an exponential quadratic kernel with a length scale $\ell=3$ and an amplitude $\sigma=1$. Each line is formed by 4000 samples. (Right) Visual representation of the kernel matrix for the samples. The diagonal indicates the variances of the noise terms and the blue oblique region implies a strong correlation between the neighbor points.}
\label{fig:functionsfromkernel}
\end{figure}

\begin{figure}[H]
\centering
\includegraphics[width=2in]{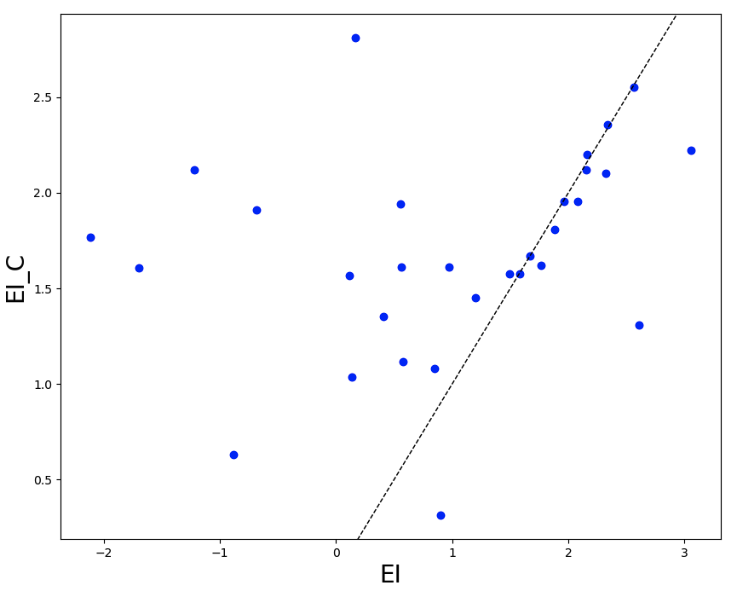}
\caption{Comparison of EI and corrected EI over $f(x_t)$ for same termination criterion $\kappa$. The dotted line represents the function $y=x$. More points are shown to be above this line, indicating that corrected EI is more likely to select the point that returns a higher value on the objective function than standard EI.}
\label{fig:profitfunctionsfromkernel}
\end{figure}

In this part, we constructed our functions based on the samples drawn from a squared exponential kernel with length scale $\ell=3$ and amplitude $\sigma=1$. As indicated by Figure~\ref{fig:functionsfromkernel} (b), the covariances between neighbor samples are not all zero given a relatively large length scale parameter. We created 30 sample sets $S_1,\cdots, S_{30}$ of 4000 data points from this kernel function as shown in Figure~\ref{fig:functionsfromkernel}(a). For each sample set $S_i$, the function $f_i$ is defined as $f_i(x)=f(x_j)+\varepsilon $ where $x_j =\arg\min_{x_j \in S_i}{\|x-x_j\|}$ and the observation noise $\varepsilon$ is Gaussian distributed with mean 0 and standard deviation $\upsilon =0.16$. We deployed BO with EI or Corrected EI to optimize these functions and the kernel was set to be the same as the sampled kernel. The performance of the acquisition function is evaluated through $f(x_t)$ corresponding to the same $\kappa$ that equals 1 percent of the maximum difference over five samples. A two-sided Wilcoxon sign rank test is performed to test the null hypothesis that the corrected EI is not different from EI under our noisy settings. The test gave a p-value equal to 0.013, indicating we should reject our null hypothesis. The scatter plot as shown in Figure~\ref{fig:profitfunctionsfromkernel} also indicates this fact.

\section{Supplements for benchmark results}

\begin{figure}[H]
\begin{minipage}[H]{0.23\linewidth}
\centering
\includegraphics[width=1.3in]{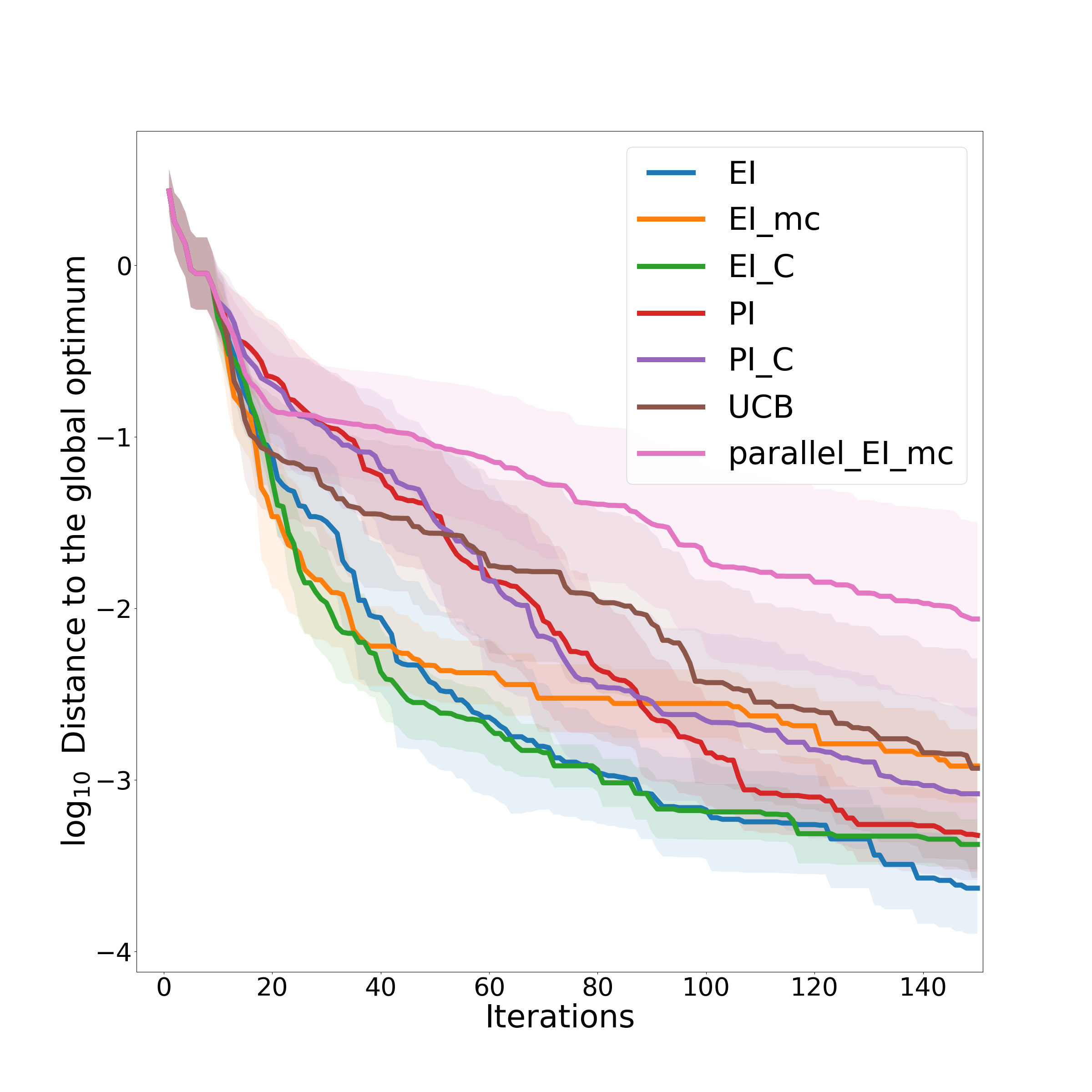}
\caption*{(a) \textit{Hartmann3d}}
\end{minipage}%
\begin{minipage}[H]{0.23\linewidth}
\centering
\includegraphics[width=1.3in]{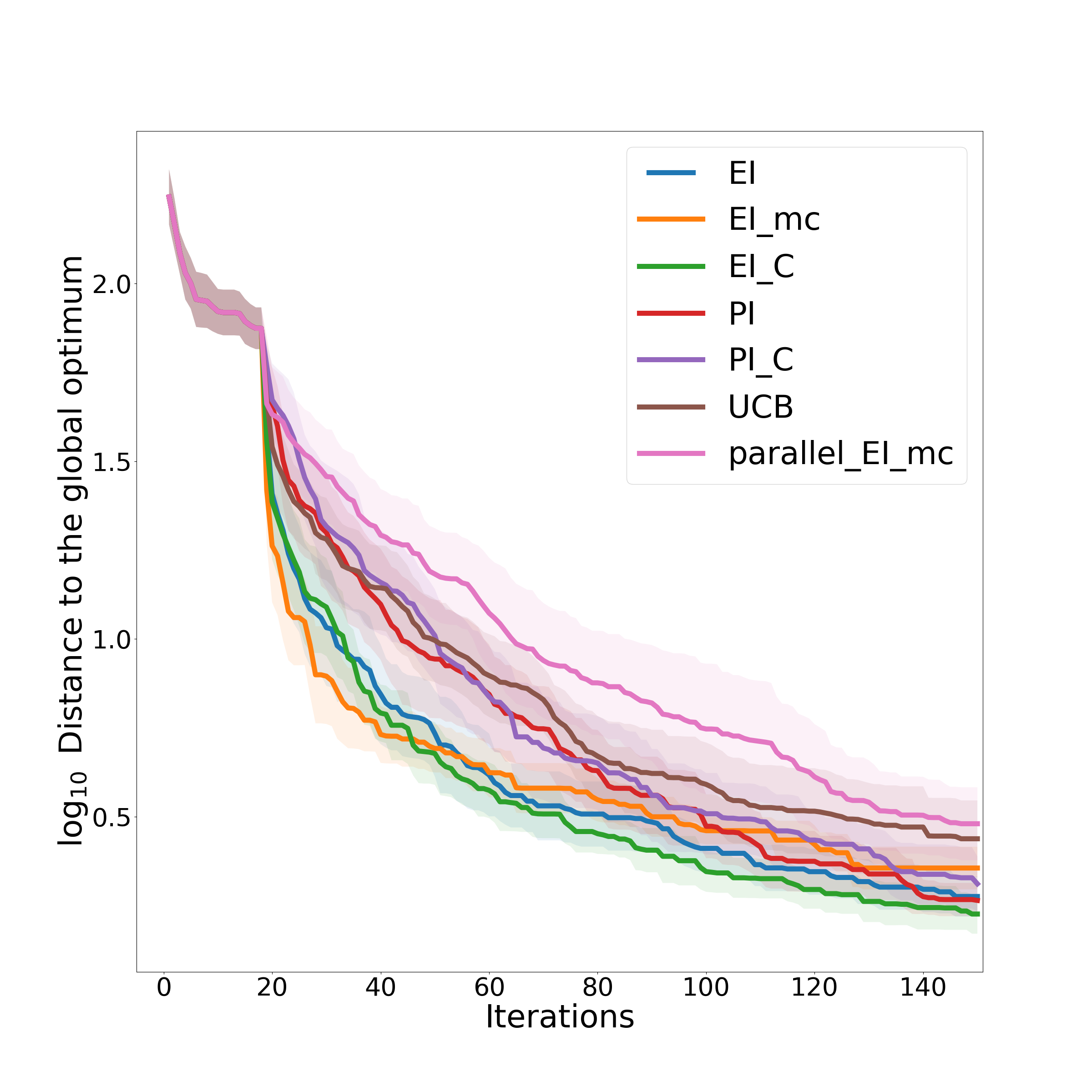}
\caption*{(b) \textit{Griewank}($d=6$)}
\end{minipage}
\begin{minipage}[H]{0.23\linewidth}
\centering
\includegraphics[width=1.3in]{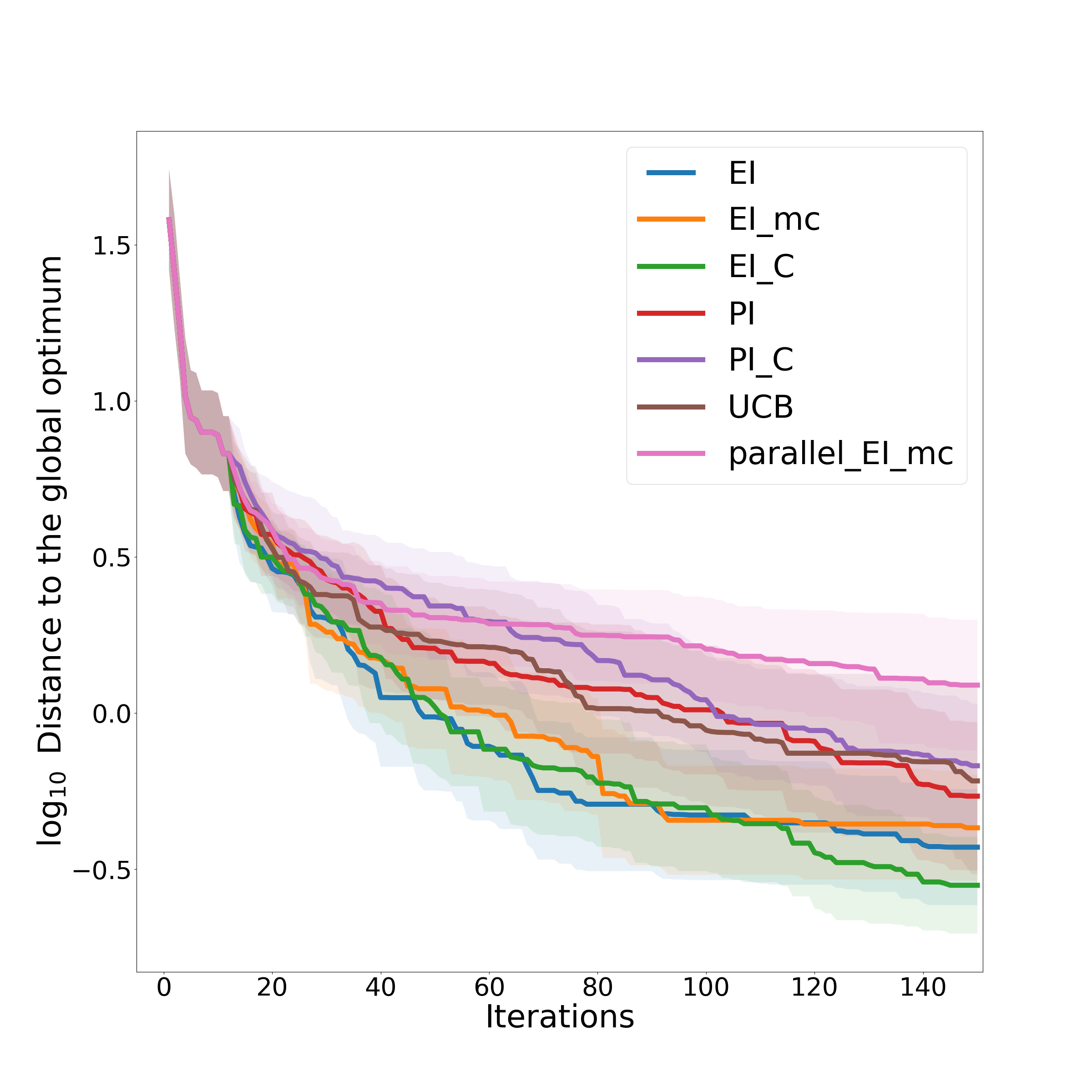}
\caption*{(c) \textit{Levy}($d=4$)}
\end{minipage}
\begin{minipage}[H]{0.23\linewidth}
\centering
\includegraphics[width=1.3in]{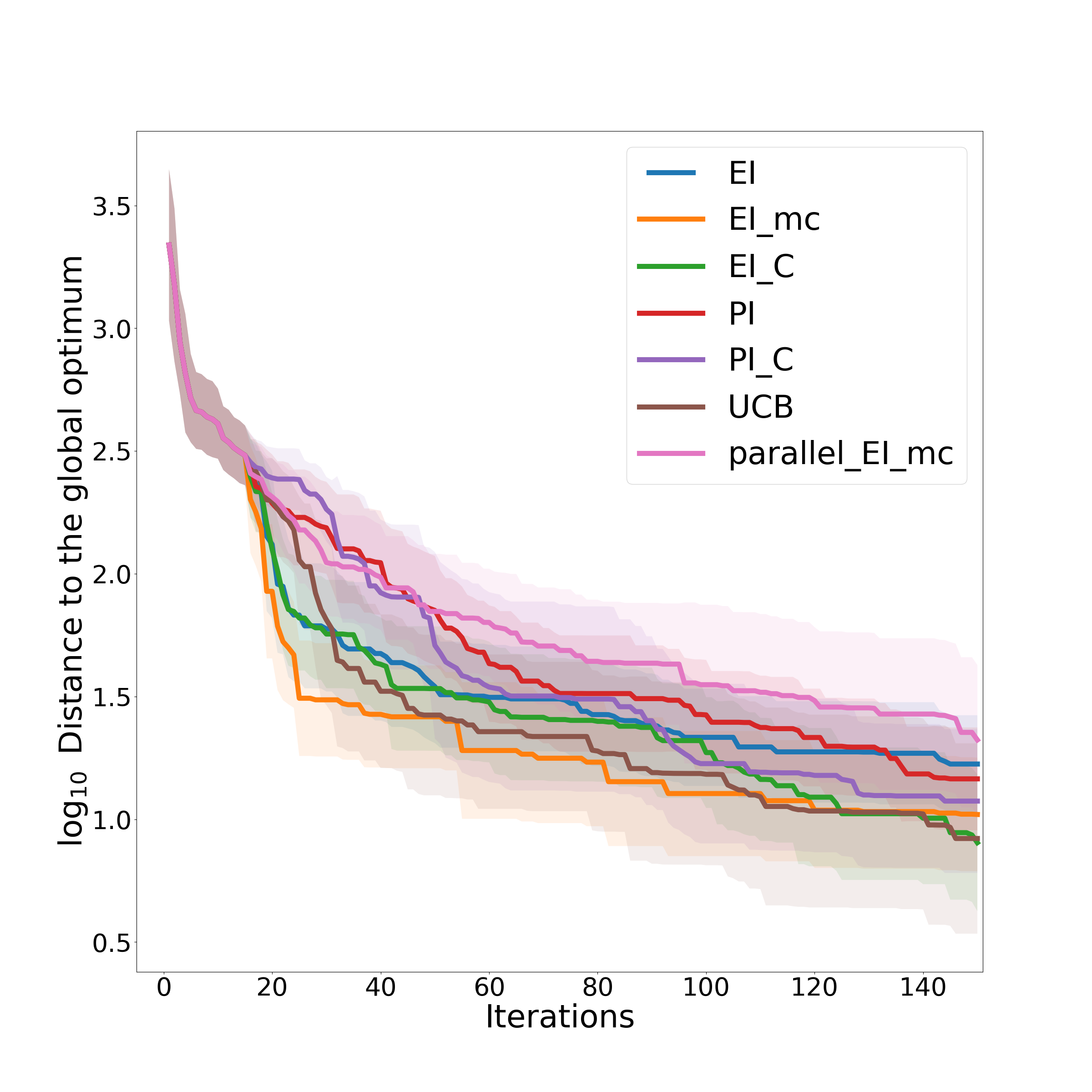}
\caption*{(d) \textit{Powell}($d=5$)}
\end{minipage}
\\
\begin{minipage}[H]{0.23\linewidth}
\centering
\includegraphics[width=1.3in]{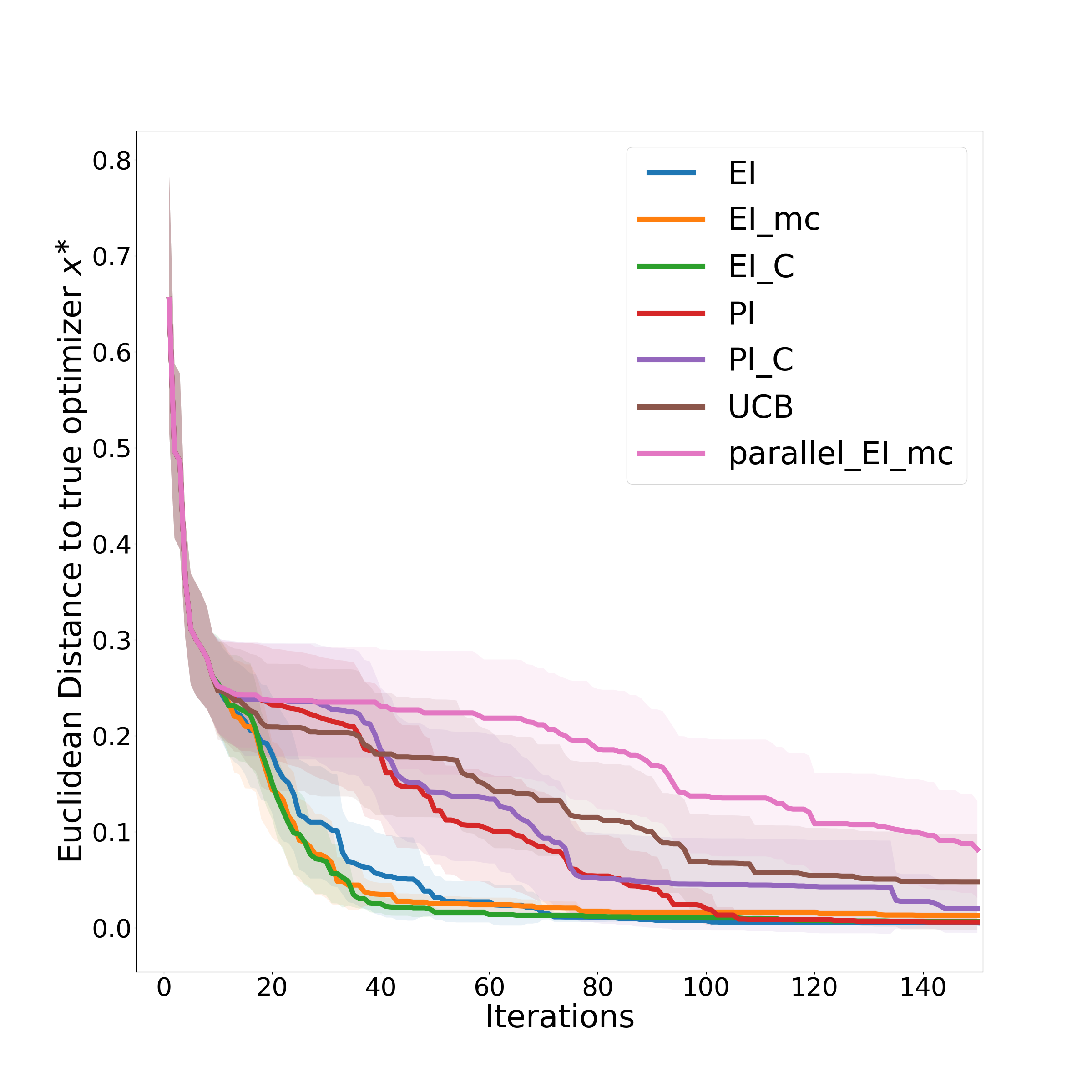}
\caption*{(e) \textit{Hartmann3d}}
\end{minipage}
\begin{minipage}[H]{0.23\linewidth}
\centering
\includegraphics[width=1.3in]{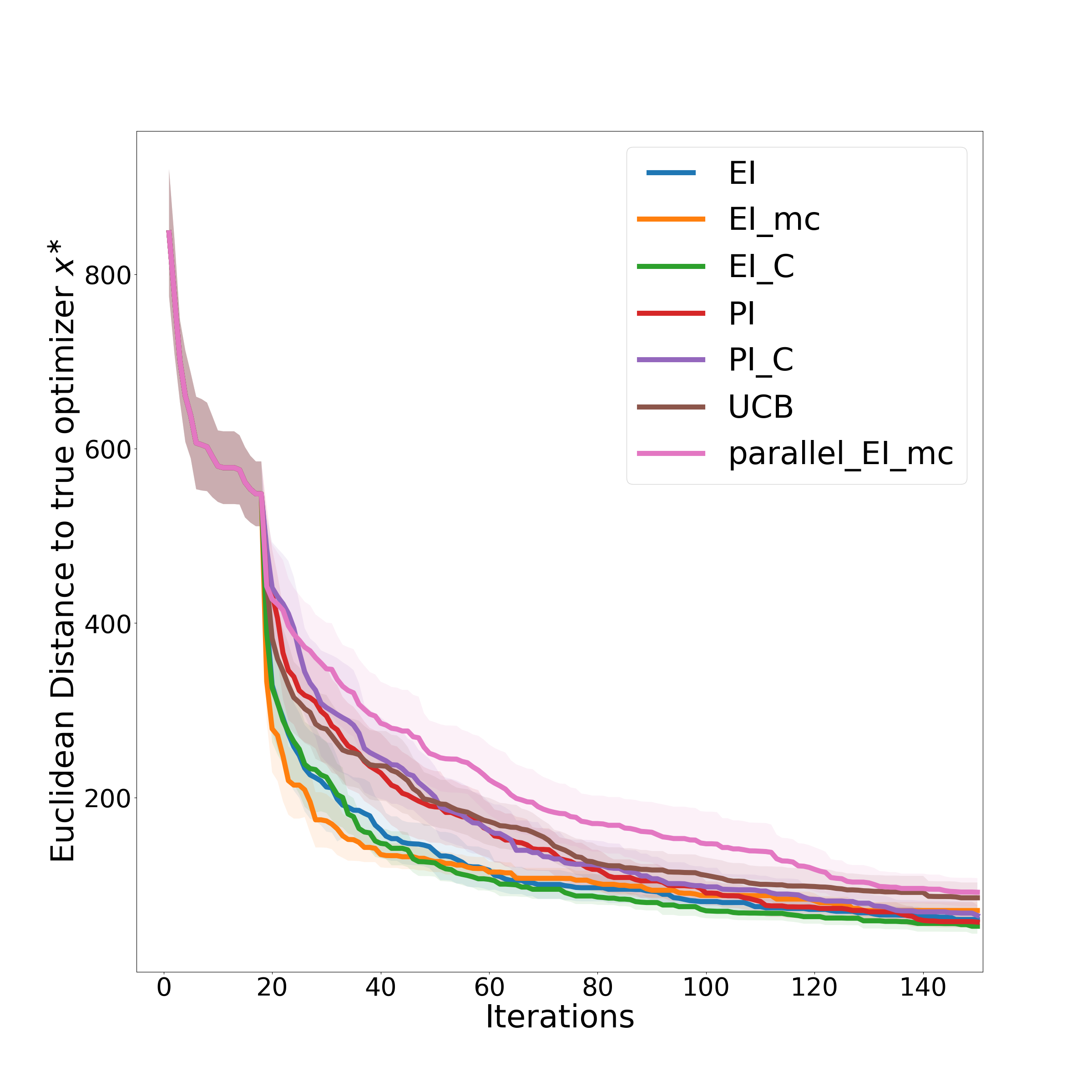}
\caption*{(f) \textit{Griewank}($d=6$)}
\end{minipage}
\begin{minipage}[H]{0.23\linewidth}
\centering
\includegraphics[width=1.3in]{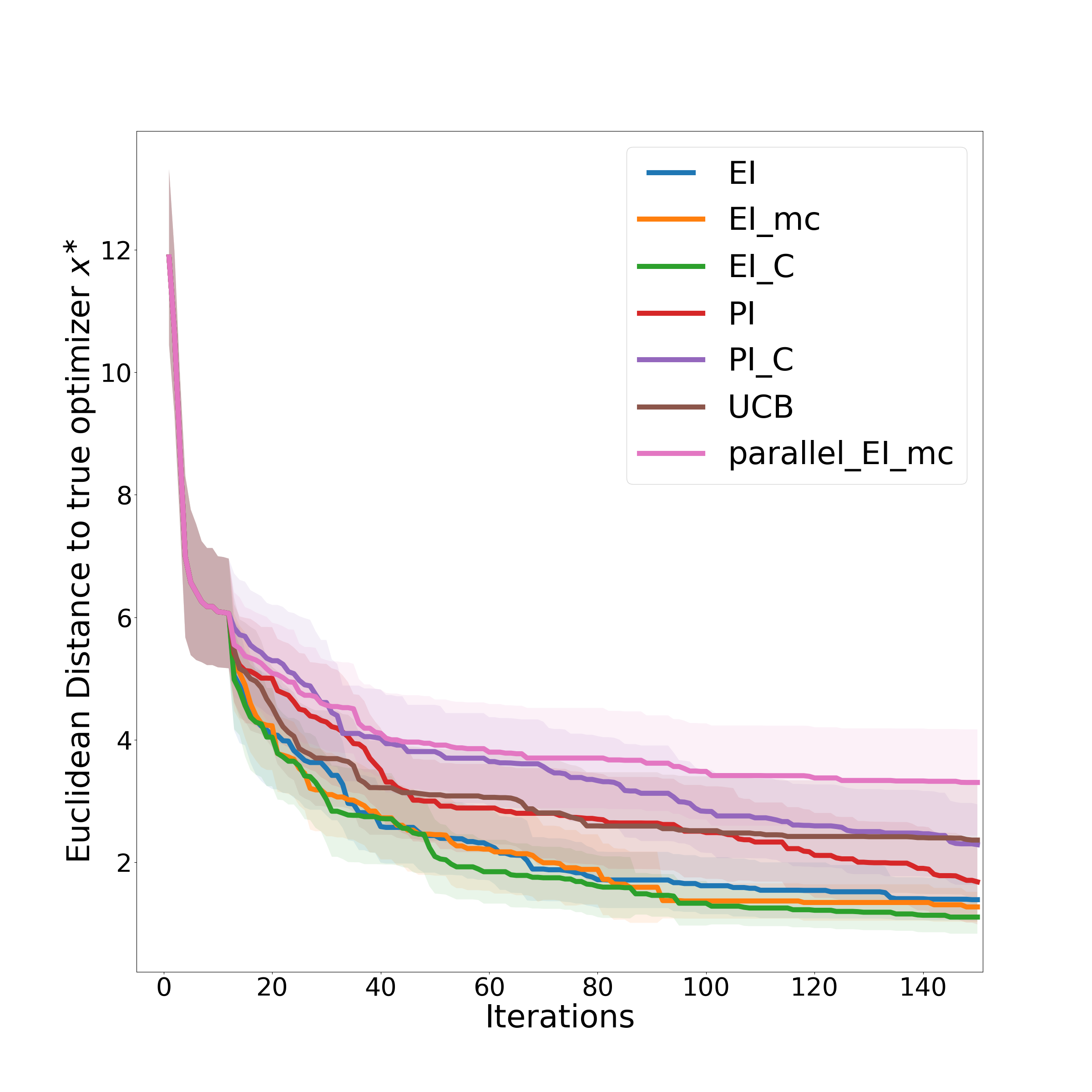}
\caption*{(g) \textit{Levy}($d=4$)}
\end{minipage}
\begin{minipage}[H]{0.23\linewidth}
\centering
\includegraphics[width=1.3in]{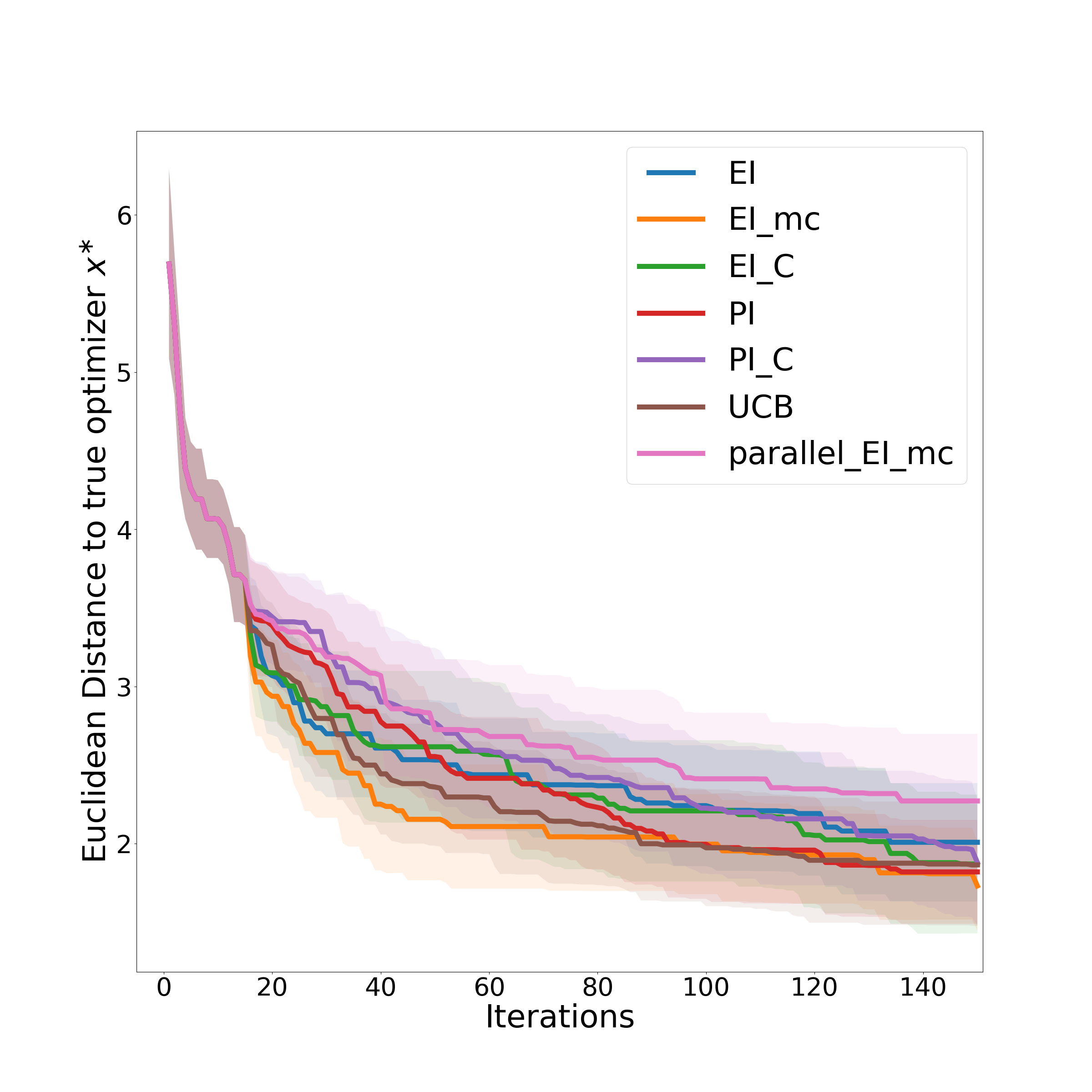}
\caption*{(h) \textit{Powell}($d=5$)}
\end{minipage}
\caption{Comparison of methods for Benchmark objective functions under the case that observation noise standard deviation $\upsilon_t$ is less than or equal to 15\% of the range of the objective function. Figures (a)$\sim$(d) show how the mean and 95\% confidence bound (shaded region) of the distance between the best feasible objective and the global optimum changes with each iteration of optimization. Figures (e)$\sim$(f) visualize the variation of the $L_2$ distance between the best point and the global optimizer $x^{\ast}$.}
\label{fig:synthetic2}
\end{figure}

\begin{figure}[H]
\begin{minipage}[H]{0.23\linewidth}
\centering
\includegraphics[width=1.3in]{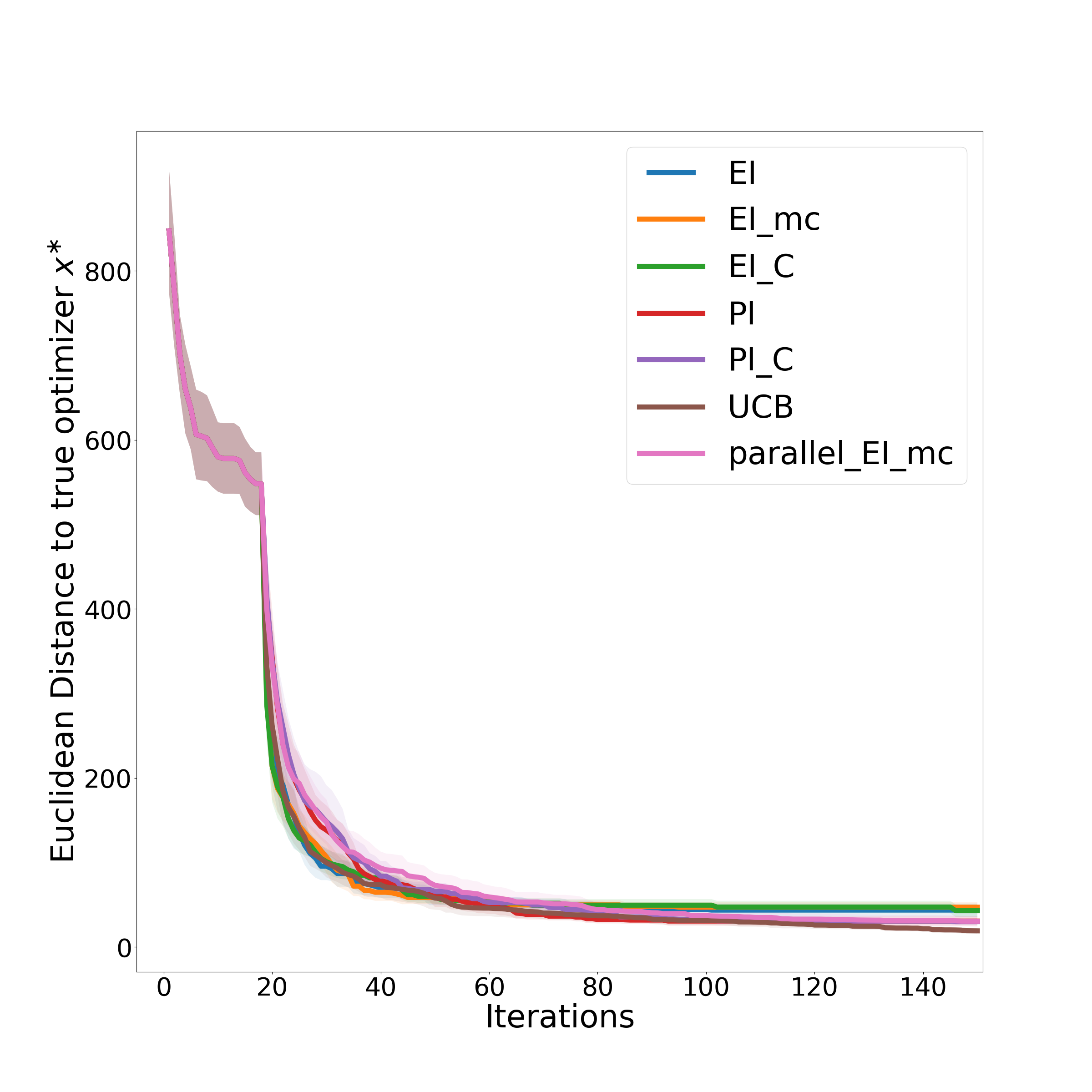}
\caption*{(a) $\upsilon_t \le 1\%$}
\end{minipage}%
\begin{minipage}[H]{0.23\linewidth}
\centering
\includegraphics[width=1.3in]{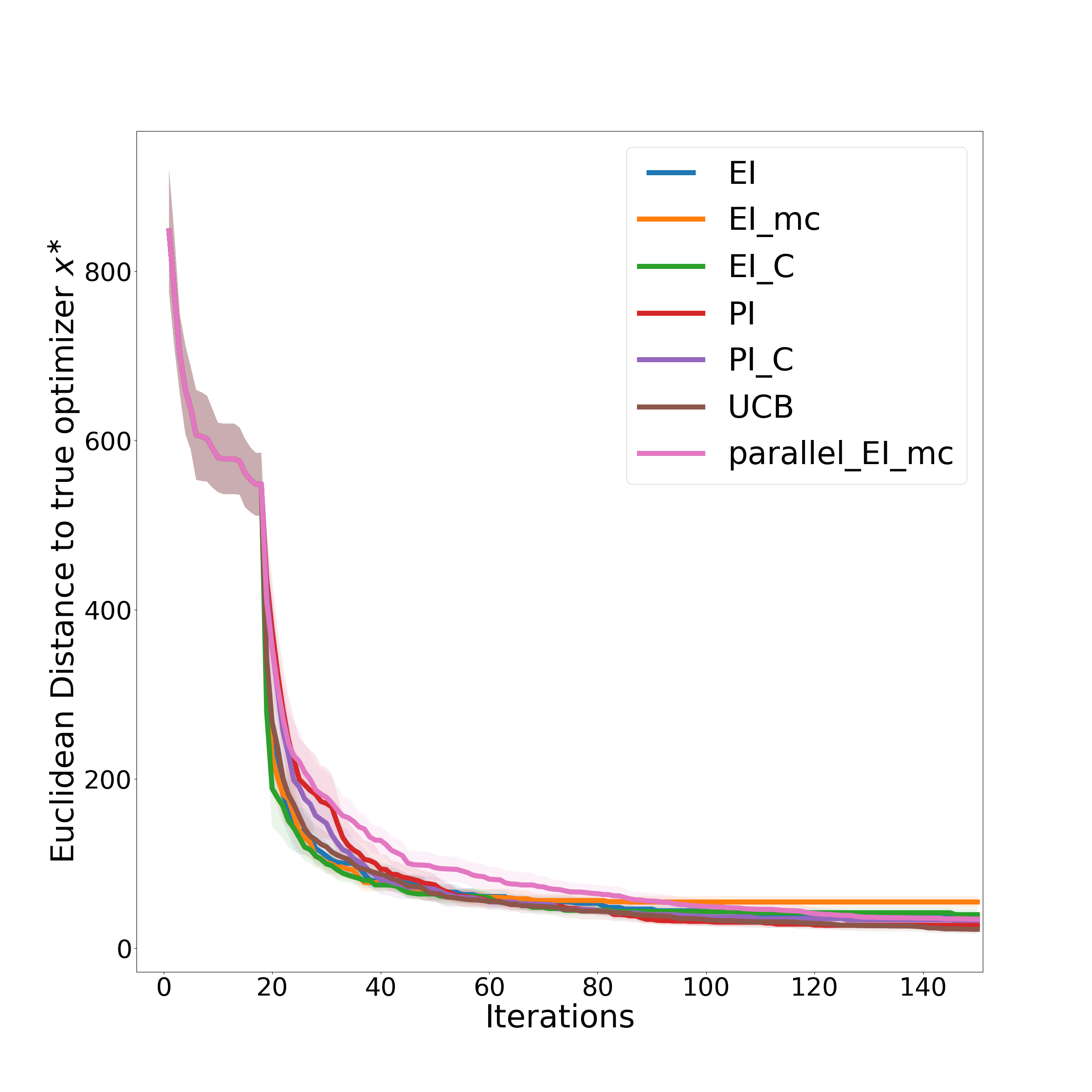}
\caption*{(b) $\upsilon_t \le 2\%$}
\end{minipage}
\begin{minipage}[H]{0.23\linewidth}
\centering
\includegraphics[width=1.3in]{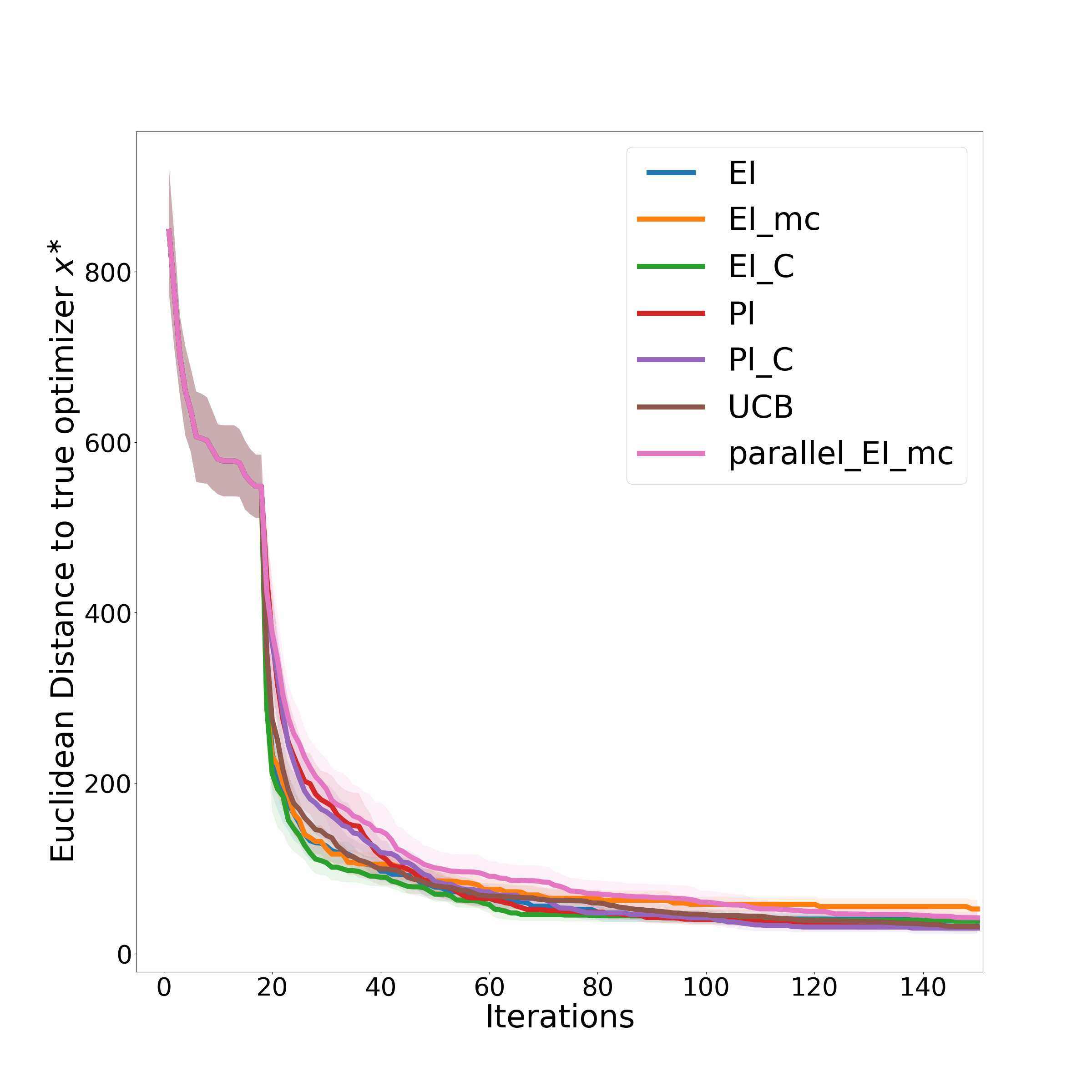}
\caption*{(c) $\upsilon_t \le 3\%$}
\end{minipage}
\begin{minipage}[H]{0.23\linewidth}
\centering
\includegraphics[width=1.3in]{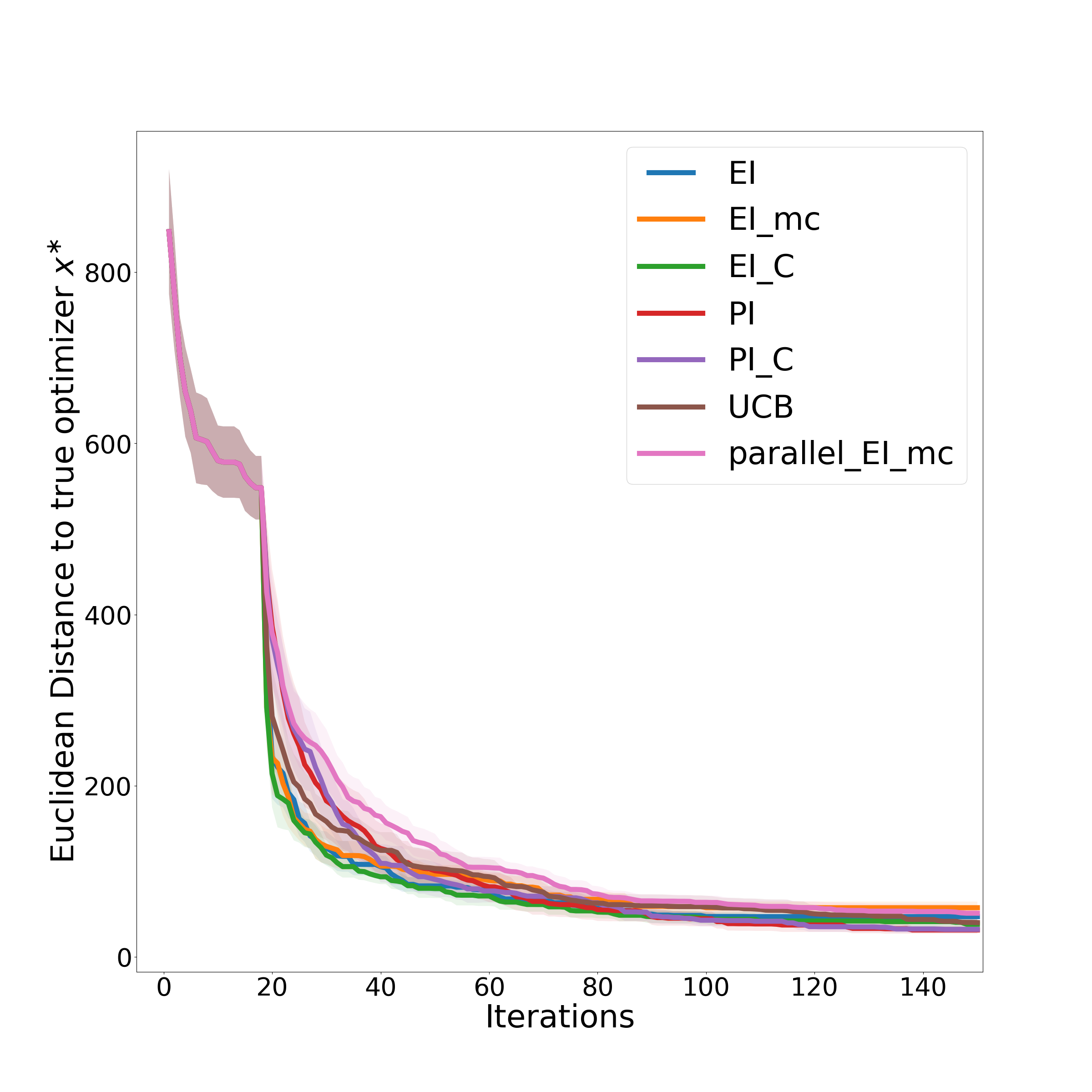}
\caption*{(d) $\upsilon_t \le 4\%$}
\end{minipage}
\\
\begin{minipage}[H]{0.23\linewidth}
\centering
\includegraphics[width=1.3in]{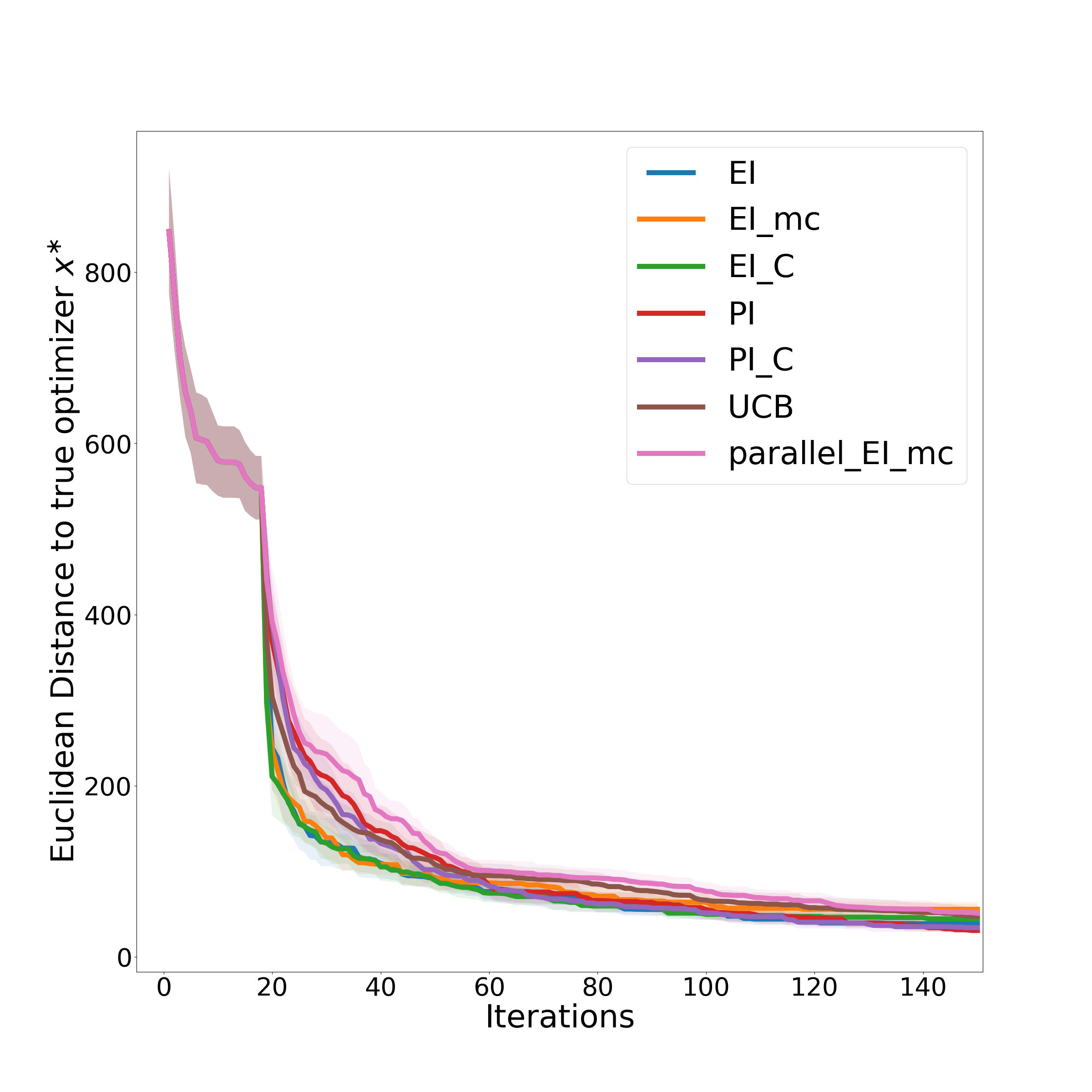}
\caption*{(e) $\upsilon_t \le 5\%$}
\end{minipage}
\begin{minipage}[H]{0.23\linewidth}
\centering
\includegraphics[width=1.3in]{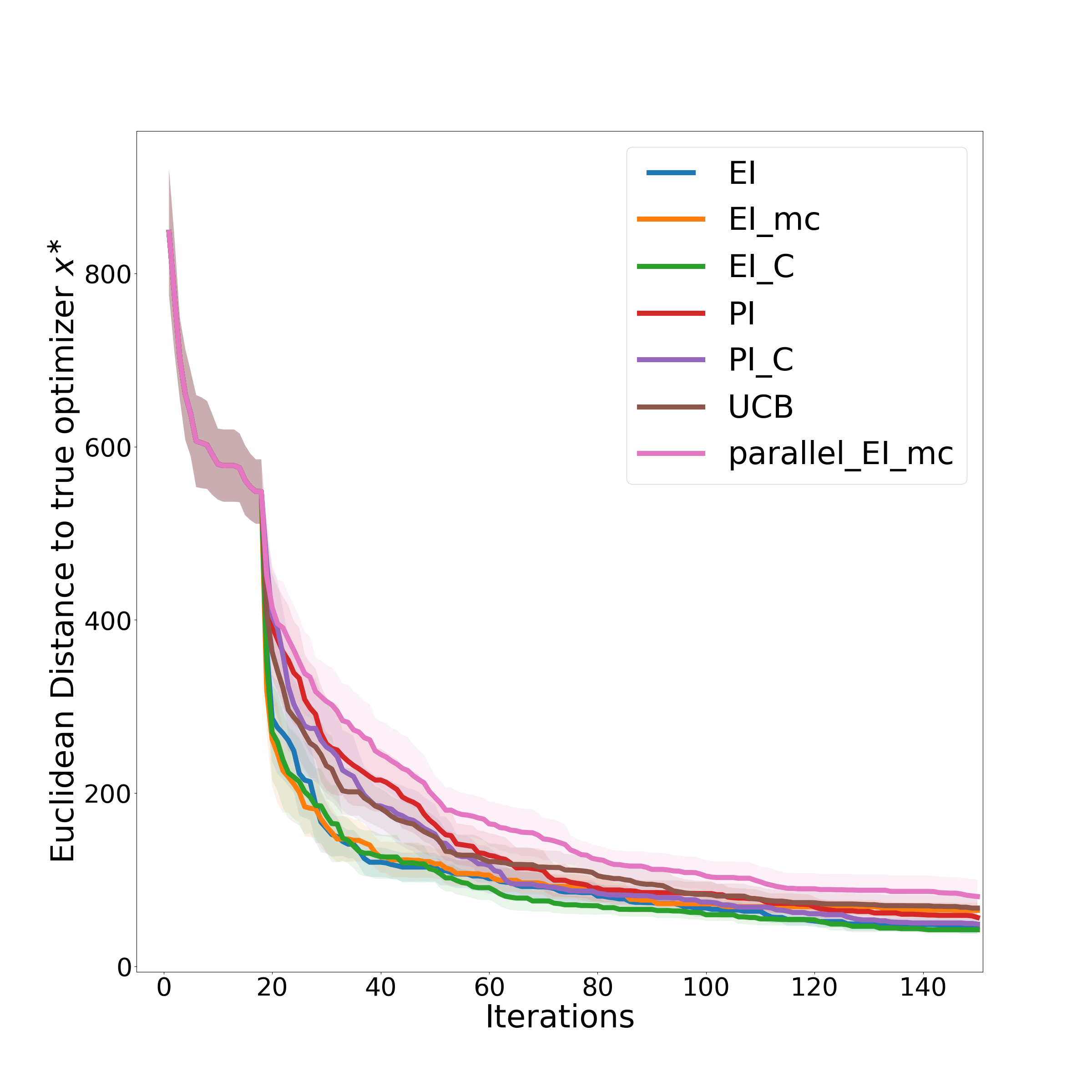}
\caption*{(f) $\upsilon_t \le 10\%$}
\end{minipage}
\begin{minipage}[H]{0.23\linewidth}
\centering
\includegraphics[width=1.3in]{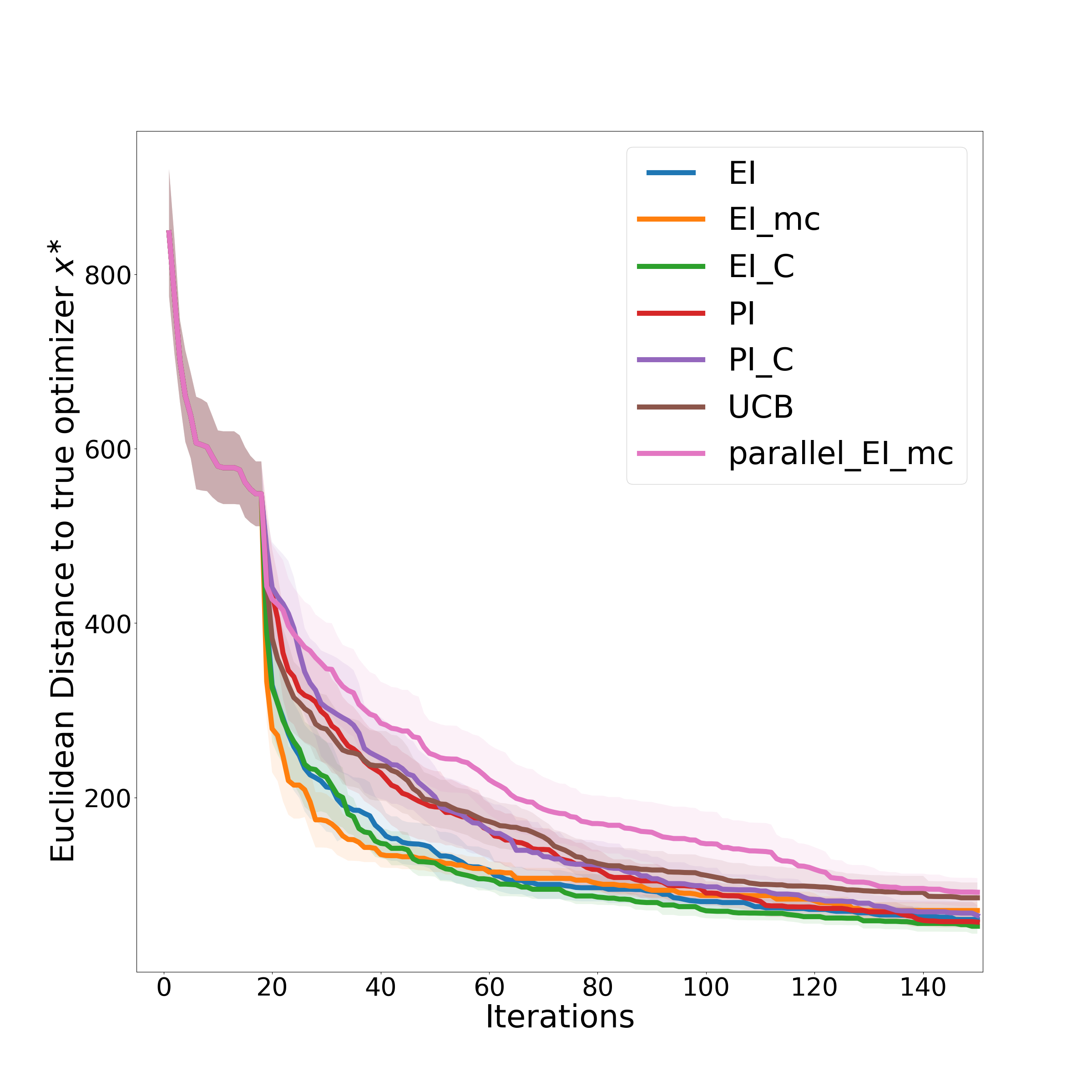}
\caption*{(g) $\upsilon_t \le 15\%$}
\end{minipage}
\begin{minipage}[H]{0.23\linewidth}
\centering
\includegraphics[width=1.3in]{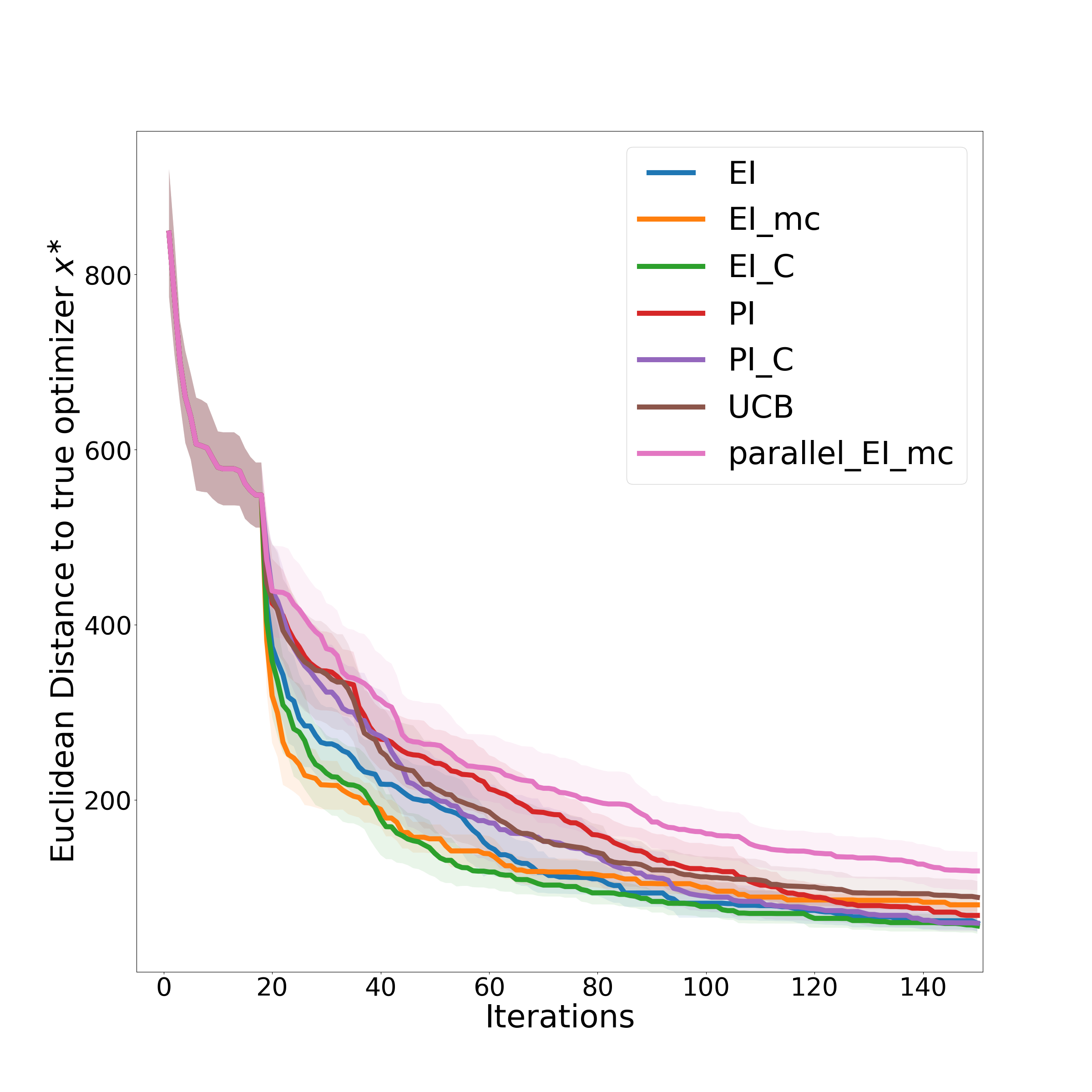}
\caption*{(h) $\upsilon_t \le 20\%$}
\end{minipage}
\caption{Optimization performance under increasing noise levels $\upsilon_t$ on a \textit{Griewank}($d=6$) function. We define the noise level as a percentage of the range of the objective function and evaluate performance by measuring the $L_2$ distance between the best point and the global optimizer $x^{\ast}$.}
\label{fig:synthetic3}
\end{figure}

We present additional results in Figure~\ref{fig:synthetic2} and Figure~\ref{fig:synthetic3}. Figure~\ref{fig:synthetic2} shows the optimization performance when the noise standard deviation $\upsilon_t$ is less than or equal to 15\% of the range of the objective function. Notably, our proposed method shows competitive performance compared to other acquisition functions. Figure~\ref{fig:synthetic3} shows the sequential optimization performance of our proposed method relative to other acquisition functions under increasing noise levels. We observe that all methods experience a decline in performance as the noise level increases, however, the corrected EI exhibits excellent performance relative to EI even in the high-noise regime.

\end{document}